\documentclass[MS]{macro/neu_msthesis}

\usepackage{physics}
\usepackage{siunitx}
\usepackage{graphics} % for pdf, bitmapped graphics files
\usepackage{graphicx} % for epx
\usepackage{epsfig} % for postscript graphics files
\usepackage{lipsum} 
\usepackage{amsmath, bm} % assumes amsmath package installed
\usepackage{amssymb}  % assumes amsmath package installed
\usepackage{upgreek}
\usepackage{url}
\usepackage{breakurl}
\usepackage{caption}
\usepackage{subcaption}
\usepackage{pgfplots}
\usepackage{floatrow}
\usepackage{float}
\usepackage[ruled,linesnumbered]{algorithm2e}
\usepackage{multicol}
\usepackage{tikz}
\usepackage{booktabs}
\usepackage{longtable}
\usepackage{algpseudocode}
\usetikzlibrary{shapes, arrows}
\usepackage{amsmath}
\newfloatcommand{capbtabbox}{table}[][\FBwidth]

\title{Analysis of Harpy's Constrained Trotting and Jumping Maneuver}

\author{Prathima Ananda Kumar}

\dept{Electrical and Computer Engineering}

\degree{Thesis}

\degreename{Masters in Robotics}

\field{Robotics}

\submitdate{August 2025}

\numberofmembers{2}

\principaladviser{Dr. Alireza Ramezani}
\firstreader{Dr.xxx}
\chairman{Dr. xxx}

\dean{Dr. xxx}

\usepackage{amsfonts}			% special math characters
\usepackage{times}		% use Postscript fonts

\newcommand{\ifno}[1]{}

\usepackage{multirow}
\makeindex
\usepackage{makeidx}
\usepackage{acronym}

\usepackage{url}
\usepackage{graphicx}
\ifnum\pdfoutput>0
\usepackage[pdftex,%                    % hyper-references for ps2pdf
bookmarks=true,%                        % generate bookmarks ...
bookmarksnumbered=true,%                % ... with numbers
hypertexnames=false,%                   % needed for correct links to figures
breaklinks=true           %Allows link text to break across lines;
]{hyperref}
\else
\usepackage[hypertex,%                  % hyper-references for ps2pdf
bookmarks=true,%                        % generate bookmarks ...
bookmarksnumbered=true,%                % ... with numbers
hypertexnames=false,%                   % needed for correct links to figures
breaklinks=true           %Allows link text to break across lines;
]{hyperref}
\fi

\hypersetup{				% setup PDF information fields
pdfauthor = {\authorRef},
pdftitle = {\titleRef},
pdfsubject = {\expandafter{\degreeRef} thesis submitted to Northeastern University},
pdfkeywords = {Multimodal robots,M4,Traversability,3D path planning}
pdfcreator = {LaTeX with hyperref package},
pdfproducer = {dvips + ps2pdf}}

\begin{document}

% add a pdf bookmark to the cover page
\pdfbookmark[1]{Cover}{cover}

% --- title page ---
\titlepage

% --- front matter ---
\begin{frontmatter}
% print signature page
%\signaturepage
% dedication
%\input{tex/dedication.tex}

% table of content (add bookmark for convenience)
\pdfbookmark[1]{Table of Contents}{contents}
\tableofcontents
\listoffigures
\newpage\ssp
%\listoftables

% include a list of Acronyms (comment out if no acronyms are specified)
% acronyms.tex

\chapter*{List of Acronyms}
\addcontentsline{toc}{chapter}{List of Acronyms}

% below is the list of acronym definitions, place them in alphabetical order
% since they will not be sorted again. 
\begin{acronym}
\acro{CAD}{Computer-Aided Design}.
The use of computer software to create, modify, analyze, or optimize engineering designs represented as precise digital models.

\acro{Dof}{Degree-of-freedom}.
The total number of independent axes or directions of motion a robot has

\acro{EoM}{Equation of Motion}.
Mathematical expressions that describe a robot’s dynamics, relating forces/torques to accelerations and motion.

\acro{DAE}{differential-algebraic equations}.
Hybrid systems of differential and algebraic constraints often used to model constrained robotic dynamics, such as closed-chain mechanisms or contact dynamics in legged robots.

\acro{EKF}{Extended Kalman Filter}
A state-estimation algorithm for nonlinear systems that linearizes around the current estimate; in robotics it is widely used for sensor fusion (e.g., IMU + vision) to estimate position, velocity, or joint states.

\acro{UDoF}{Underactuated degrees-of-freedom}.
Motion dimensions not directly actuated by motors but instead influenced by passive dynamics; common in legged robots where certain body or joint motions are indirectly controlled.

\acro{PID}{Proportional-Integral-Derivative}.
A feedback control law using proportional, integral, and derivative terms; in robotics it regulates joint positions, velocities, or forces, and is especially common in low-level motor controllers.

\acro{COM}{Center of Mass}.
The effective point at which the total mass of the robot is concentrated; critical in locomotion control since COM trajectory determines stability, balance, and jumping dynamics.

\acro{MPC}{Model Predictive Control}.
An optimization-based control framework that predicts future robot states over a horizon using a dynamics model, and computes control inputs (e.g., ground reaction forces, torques) subject to physical constraints, often used in bipedal/quadruped locomotion.

\acro{VMC}Virtual Model Control.
A control method that applies “virtual forces and springs” in simulation space to generate desired behaviors; in robotics it is used to design intuitive controllers for leg compliance, balance, and hopping height regulation.

\acro{EDF}{Electric ducted fan}.
  A model of a system in which communication is described as transactions, abstract of pins and wires.
  In addition to what is provided by the, it models arbitration on a bus transaction level.

\acro{PWM}{Pulse width modulation}.
    System bus definition within the AMBA 2.0 specification. Defines a high-performance bus including pipelined access, bursts, split and retry operations.
\end{acronym}

% include any of the front matter files that contain text
% attention the input does cause a page break, the include on 
% the other hand does not
% acknowledgements.tex:

\begin{acknowledgements}

The successful completion of this thesis was made possible by the invaluable support and guidance of the remarkable individuals at SiliconSynapse Lab at Northeastern University. I would like to express my deepest gratitude to my advisor, Prof. Alireza Ramezani for constant support, guidance, encouragement, and enlightenment throughout my journey as a Thesis student and instruction during Legged robotics course. His vision and enthusiasm for legged robotics inspired me to choose this field as the focus of my career. I am equally grateful to my PhD mentor, Shreyansh Pitroda, for their insightful advice, patience, and mentorship, which greatly enriched my academic growth. I wish to thank my fellow researchers: Bibek Gupta, Sushma Aare, Darshan Krishnamoorthy, Kaushik Venkatesh, Adarsh Salagame  for their assistance, collaboration and encouragement. 

I extend my sincere gratitude to my Thesis committee members, Prof. Hanumant Singh and Prof. Derya Aksaray for their insightful discussions, feedbacks during my thesis defense and for the guidance I received in their courses. I greatly admire their expertise in their respective fields and enjoyed the opportunity to learn under their instruction.

Atlast, I am forever grateful to my father, mother, sister and my family for both my personal and career growth. I would like to dedicate my masters degree to my family.

\end{acknowledgements}

% abstract.tex:

\begin{abstract}
This study presents an analysis of experimental data from Harpy, a thruster-assisted bipedal robot developed at Northeastern University. The study examines data sets from trotting and jumping experiments to understand the fundamental principles governing hybrid leg-thruster locomotion. Through data analysis across multiple locomotion modes, this research reveals that Harpy achieves stable locomotion with bounded trajectories and consistent foot placement through strategic leg-thruster synergy. The results demonstrate controlled joint behavior with low torques and symmetric tracking, accurate foot placement within kinematic constraints despite phase-transition perturbations, and underactuated degree-of-freedom stability without divergence. Energy level analysis reveal that legs provide primary propulsion, while the thrusters enable additional aerial phase control. The analysis identifies critical body-leg coupling dynamics during aerial phases that require phase-specific control strategies. Consistent repeatability and symmetry across experiments validate the robustness of the hybrid actuation approach.
\end{abstract}

\end{frontmatter}

% --- body of the document ---

%\pagestyle{plain}
\pagestyle{headings}

% include each chapter like below

\chapter{Introduction}
\label{chap:intro}

The pursuit of versatile and robust locomotion in legged robots has been a central challenge in robotics research. Legged systems are capable of navigating unstructured environments, overcoming obstacles, and maintaining mobility where wheeled or tracked platforms fail. However, legs alone face inherent trade-offs between efficiency, stability, and agility, particularly when rapid transitions, large jumps, or steep terrains are involved. To address these challenges, recent research has explored multimodal locomotion, in which legged motion is combined with aerial or thruster-assisted behaviors to expand mobility capabilities. Such hybrid designs draw inspiration from nature, where many animals seamlessly combine terrestrial and aerial strategies to enhance survival.

In this thesis, we focus on Harpy~\cite{pitroda_capture_2024,pitroda_conjugate_2024,sihite_posture_2024,pitroda_quadratic_2024}, a bipedal robot equipped with a torso-mounted thruster. Harpy was developed at Northeastern University with the goal of investigating the integration of leg dynamics and aerial assistance to enable advanced maneuvers such as high jumps and slope traversal. The specific scope of this work involves a static analysis of trotting and jumping behaviors, examining how ground reaction forces and thruster contributions interact to enable effective lift-off and landing strategies. 

The design of Harpy~\cite{dangol_control_2021,dangol_hzd-based_2021,dangol_performance_2020,dangol_reduced-order-model-based_2021, sihite_unilateral_2021, de_oliveira_thruster-assisted_2020} is inspired by the widowbird, a bird species that performs exaggerated, high-energy vertical jumps using both its wings and legs. The widowbird’s strategy highlights the value of combining leg power with aerial augmentation to achieve maneuvers that would be difficult for either mode alone. By embedding a thruster at the torso, Harpy embodies this biological principle in engineered form, allowing investigation into how thrust can complement leg mechanics to extend the performance envelope of biped robots.

Harpy~\cite{pitroda_dynamic_2023,dangol_thruster-assisted_2020,pitroda_enhanced_2024} is part of a broader line of multi-modal robots developed in our lab, each exploring unique ways of combining legs and aerial propulsion. For instance, Husky~\cite{krishnamurthy_enabling_2024,krishnamurthy_narrow-path_2024,krishnamurthy_optimization_2024,krishnamurthy_thruster-assisted_2024,krishnamurthy_towards_nodate,ramezani_generative_2021,sihite_actuation_2023,ramezani_morpho-functional_2023,salagame_quadrupedal_2023,sihite_optimization-free_2021} is a quadruped robot with a quadcopter mounted on its torso. This configuration allows it to exploit both legged stability and aerial thrust for enhanced mobility in cluttered or irregular environments. Building upon this idea, Husky-$\beta$ advances the concept further by integrating individual propellers at each knee, enabling fine-grained control of thrust distribution across the body and legs. This design emphasizes localized thrust assistance for stabilization, load sharing, and agile maneuvering. Another example is the M4~\cite{sihite_multi-modal_2023,mandralis_minimum_2023,gherold_self-supervised_2024,sihite_dynamic_2024} robot, which embodies modularity and multimodality by combining wheels, legs, and aerial propulsion. M4~\cite{sihite_efficient_2022,sihite_demonstrating_2023}demonstrates the potential of integrated hybrid platforms to navigate terrains that are otherwise inaccessible to traditional locomotion strategies.

While quadrupeds like Husky and Husky-$\beta$~\cite{wang_legged_2023,wang_quadratic_2025} demonstrate the potential of distributed propulsion in multi-legged robots, Harpy emphasizes the bipedal case, where balance, thrust-vectoring, and leg coordination are more tightly coupled. Static analysis of Harpy’s trotting and jumping behaviors is therefore an essential first step in quantifying the mechanical interactions between thrust and leg dynamics in bipeds, and in laying the groundwork for dynamic controllers that exploit this synergy.

The motivation for this thesis stems from the need to build deeper insight into Harpy’s design and performance capabilities. Specifically, the research aims to understand how the robot achieves self-sustained operation, how well it maintains robust performance under high-impact events such as landing impacts, how effectively it coordinates leg–body interaction during ballistic phases, and to what extent the thrusters contribute to achieving higher jumps. These insights are essential to developing future control strategies that ensure both stability and efficiency in dynamic maneuvers.

Through these analyses, the thesis aims to establish a foundation of experimental evidence and data-driven insights that not only validate Harpy’s design but also guide the development of more advanced controllers capable of handling dynamic, underactuated, and high-impact robotic behaviors.

\subsection{Thesis Objectives}
The objectives of this work are organized around evaluating Harpy’s performance within a constrained gantry setup. The primary goals are as follows:
\begin{itemize}
    \item Evaluate the performance of the gantry system, with particular emphasis on verifying that the imposed constraints are maintained and that sagittal-plane dynamics are effectively isolated.  
    \item Assess Harpy's performance across three key aspects: body position and orientation tracking, foot placement accuracy, and joint-level tracking. These metrics are critical for validating the effectiveness of the locomotion control framework.  
    \item Investigate the stability of underactuated coordinates during both trotting and jumping maneuvers.  
    \item Formulate an energy-based argument to quantify the relative contributions of the legs and thrusters during jumping.  
    \item Examine the influence of leg motion on overall body dynamics throughout jumping experiments.  
\end{itemize}

\subsection{Thesis Contributions}
This thesis makes three primary contributions:  
\begin{itemize}
    \item Partial involvement in experimental data collection for both trotting and jumping trials, ensuring the availability of high-quality datasets for subsequent analysis.  
    \item Execution of detailed statistical analyses on the collected datasets to identify key performance patterns, trends, and deviations.  
    \item Derivation of critical observations and conclusions from these analyses, which directly inform and guide the next stage of controller design, thereby contributing to improved system robustness and effectiveness.  
\end{itemize}

% Lit_review
\chapter{Literature Review}
\label{chap:Literature Review}

A springtail-inspired microrobot \cite{ramirez_serrano_springtail-inspired_2025} achieves multimodal locomotion at the centimeter scale by combining cyclic legged ambulation with a torque-reversal catapult mechanism that releases stored elastic energy through a segmented appendage, enabling repeatable directional take-offs and upright landings. While relying mainly on passive mechanical dynamics, it demonstrates efficient transitions between walking and jumping. Similarly, a quadruped model with a pre-loaded elastic prismatic spine \cite{ye_modeling_2021} integrates spinal compliance with trajectory optimization, allowing stored elastic energy to amplify motor output and extend standing long jumps by 23$\%$ compared to rigid-body counterparts. In parallel, a biped robot with a hierarchical control framework \cite{zhang_design_2023} incorporates a kino-dynamic planner and a heuristic landing strategy based on momentum feedback, achieving stable execution of diverse tasks such as twisting jumps, directional leaps, and somersaults. A parallel-legged robot \cite{he_controllable_2021} instead emphasizes controllable hopping height through BLDC–harmonic drive actuation regulated by virtual model control, ensuring repeatable vertical hops with minimal sensing and extending naturally to spatial hopping scenarios. At the larger scale, the MIT Cheetah 3 quadruped \cite{nguyen_optimized_2019} demonstrates powerful platform jumps through trajectory optimization of the full-body dynamics, coordinated by high-frequency PD tracking and a force-based landing controller, with hardware validation showing reliable leaps onto and off a 0.76-meter platform. A four-wheeled robot augmented with a tail \cite{iwamoto_jumping_2015} exploits tail actuation for lift generation and aerial orientation, with trajectory planning enabling mid-air repositioning confirmed through simulation. Salto-1P \cite{haldane_repetitive_2017}, a monopedal platform with series-elastic leg actuation, thrusters, and an inertial tail, achieves extreme saltatorial locomotion with vertical leaps up to 1.25 m and continuous hopping at accelerations exceeding 14 g, coordinated by a linearized Raibert step controller and experimentally validated through over 700 jumps. Complementing these experimental platforms, a comprehensive survey of bio-inspired jumping systems \cite{mo_jumping_2020} synthesizes strategies from insects, frogs, galagos, kangaroos, and dolphins alongside robotic counterparts, comparing mechanisms for elastic energy storage, snap-through actuation, and multi-modal propulsion. The review further analyzes control strategies for take-off, aerial righting, and landing buffering, while benchmarking performance across robots and animals, and outlines future research directions in compliant design, multi-modal locomotion, and swarm coordination.

A hybrid aerial–terrestrial platform \cite{pratt_dynamic_2016} combines a quadrotor body with passive-dynamic legs, enabling passive walking down slopes and active walking on flat or inclined surfaces through rotor thrust. Control relies on dynamic modeling and simulation of coupled aerial and leg dynamics, and evaluation is conducted via treadmill trials with Vicon motion capture to validate stable limit cycles in passive walking and thrust-assisted active locomotion. A wheeled bipedal robot \cite{zhao_design_2024} adopts a squat-inspired leg design that balances torque between hip and knee actuators, paired with a height-variable wheeled LIP model and CLF-based whole-body control, achieving robust velocity tracking, squat motions, and disturbance rejection in both simulation and hardware. A multi-locomotion parallel-legged bipedal robot \cite{wang_design_2024} employs a five-link double-crank leg structure to realize walking, bouncing, and hopping; control is based on a combination of inverted pendulum modeling, virtual model control, and decoupling balance algorithms, validated through simulation and prototype hopping tests that achieved up to 0.95 m vertical jumps. At the humanoid scale, a CoP-guided angular momentum controller \cite{qi_vertical_2023} regulates vertical jumps by shaping take-off momentum and actively absorbing impact upon landing; evaluation through repeatable hardware trials shows improved stability, reduced joint load, and consistent recovery across multiple jumps.

A quadrotor-assisted biped robot \cite{maekawa_pseudo-locomotion_2018} demonstrates pseudo-locomotion by using aerial thrust for stability while its legs execute foot trajectories optimized via deep reinforcement learning. The approach generates walking-like motions with ground-contact velocity regulation and is validated in both simulation and hardware, producing agile but tethered motions. A modular torque-controlled robot architecture \cite{grimminger_open_2020} provides a standardized platform for legged locomotion research, enabling experiments in walking, running, and jumping across multiple robot morphologies; torque transparency allows for impact-rich tasks like hopping to be studied under consistent control interfaces. Extending multi-modal capabilities, a small humanoid robot \cite{sugihara_design_2023} integrates wheels and a fully-actuated trirotor flight unit, with an integrated control framework that coordinates legged, wheeled, and aerial locomotion, experimentally showing seamless transitions between modes including flight-assisted jumps and aerial manipulation. A musculoskeletal bipedal robot \cite{li_design_2024} powered by pneumatic artificial muscles is explicitly designed to emulate the spring-loaded inverted pendulum (SLIP) model; sequential jumping experiments reveal ground reaction forces, COM trajectories, and stiffness profiles that closely match SLIP simulations, validating biologically inspired design principles. At a smaller scale, a miniature humanoid platform \cite{liu_design_2022} uses proprioceptive actuation with cable-driven hips and linkage-based ankles to lower leg inertia, combined with a convex model hierarchy predictive control framework that enables push recovery, CoM tracking, and preliminary vertical jumping demonstrations, establishing an accessible system for dynamic motion research.

A quadruped robot study \cite{suzuki_foot_2025} demonstrates how variations in foot trajectory design particularly hip–knee coordination directly shape emergent gait patterns such as trot, pace, and gallop. Using a Unitree A1 platform in simulation with a CPG-based controller, the evaluation reveals that trajectory parameters affect stability, energy efficiency, and adaptability on rough terrain, establishing foot trajectory as a key determinant of locomotion performance. A jumping quadruped concept \cite{kikuchi_basic_nodate} employs spider-like legs with four-bar linkages powered by pneumatic cylinders, enabling rapid high-power take-off while also acting as shock absorbers during landing. Experiments with a one-leg prototype validate both the power output of pneumatic actuators and their damping capacity, achieving ~350 mm jumps and confirming stable impact absorption for rough-terrain mobility. At the bipedal scale, an optimized deep learning approach \cite{challa_optimized-lstm_2022} combines RGB-D sensing (Kinect V2) with an LSTM-based trajectory generator, where hyperparameters are tuned via the Rao-3 algorithm to capture human gait and reproduce stable walking. Evaluation includes gait stability metrics (cyclograms and phase plots) and simulation on the HOAP-2 humanoid, confirming the ability to generate human-like, stable lower-limb trajectories.

A broad review of animal-inspired strategies \cite{mo_jumping_2020} surveys invertebrate, vertebrate, and aquatic jumpers, and translates these mechanisms into robotic design. It emphasizes catapult-based elastic actuation, snap-through buckling, and tendon-driven power amplification, with control strategies focused on take-off angle, aerial righting, and landing buffering; evaluation is presented via comparative performance benchmarks across animals and robots. Complementing this, work on quadruped robots \cite{wang_research_2023} imitates animal posture adjustment to generate jumping motions, employing trajectory optimization and model predictive control (MPC) to track reference trajectories. Simulation trials confirm the ability to clear 40 cm obstacles, demonstrating both agility and stability in controlled environments. A musculoskeletal humanoid platform \cite{li_design_2024} integrates pneumatic artificial muscles to replicate SLIP-like dynamics, with sequential jumping trials showing COM trajectories and ground reaction forces that align with theoretical spring-mass predictions. Another approach to bipedal gait generation \cite{wang_bipedal_2024} applies Bessel interpolation to generate smooth lower-limb trajectories on the Roban humanoid, enabling stair-walking experiments; though focused on walking, the method addresses trajectory continuity issues critical to stable transitions into and out of dynamic movements like jumps. Finally, a wheel‑legged robot \cite{guo_design_2022} with a parallel four‑bar leg proposes LQR for balance and a fuzzy‑PD jump controller, and validates jumping/obstacle‑crossing via simulations and real‑scene tests in complex terrain .

% Harpy-hardware-details

\chapter{Northeastern's Harpy Platform}
\label{chap:Northeastern's Harpy Platform}
\section{Hardware Overview or Harpy Mechanical Overview}

\begin{figure}[H]
  \centering
    \includegraphics[width=0.8\textwidth]{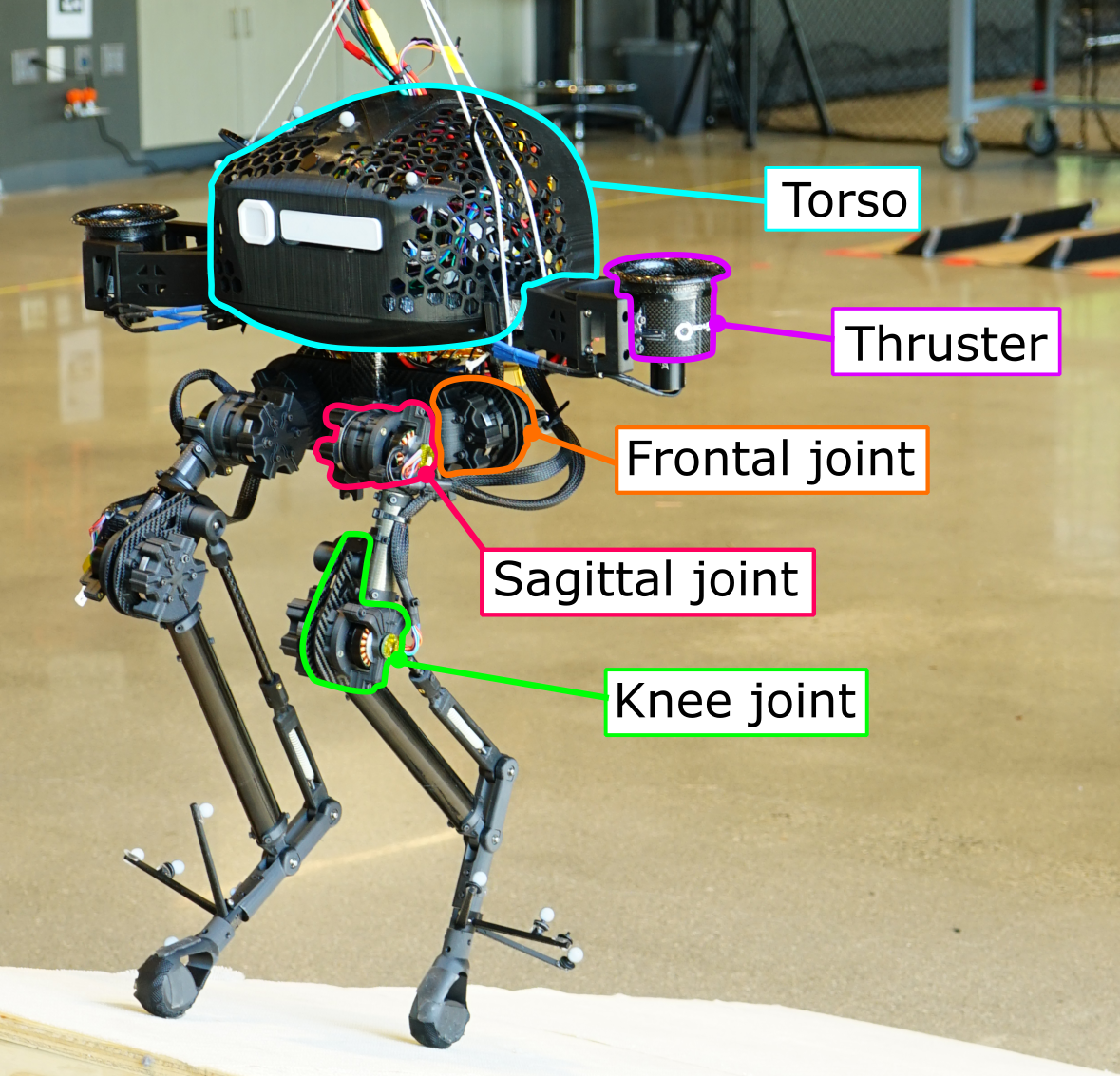}
  \caption{Harpy V2 designed and developed at Northeastern University}
  \label{fig:harpy-hardware}
\end{figure} 

Harpy shown in Fig.~\ref{fig:harpy-hardware} is a 12 Dof bipedal robot equipped with a pair of upward facing thrusters mounted on either side of its torso. The robot stands 0.7 m tall with a total mass of 6.5 kg and has a point-foot design. The thrusters are strategically positioned near the robot's CoM to reduce undesired pitch moments during operation. Harpy's joints include three per leg hip frontal, hip sagittal, and knee along with actuated degrees of freedom for body translation and rotation.

The robot's body is constructed primarily from lightweight materials, including carbon fiber tubes, carbon fiber plates, and reinforced 3D-printed components. The torso is centered around a robust yet lightweight beam comprising an aluminum honeycomb core sandwiched between two thin carbon fiber sheets, offering an optimal balance between strength and weight. This section houses all critical electronic subsystems, including micro-controllers, motor drivers, and thrusters. The legs are assembled using oval carbon fiber tubes connected via 3D-printed structurally reinforced joints. These joints are fabricated using a MarkForged 3D printer with integrated continuous carbon fiber reinforcement, significantly enhancing stiffness and mechanical strength.

Robot joint actuators are custom built using T-Motor units coupled with Harmonic Drive gearboxes, employing a 50:1 reduction ratio for all the joints. These motors are controlled via ELMO Gold Twitter servo drives, with motor commands transmitted through the EtherCAT communication protocol. Joint angles are measured using Hall effect encoders, which also facilitate tracking of desired joint positions.

The propulsion system comprises Schubeler DS-51 90mm Electric Ducted Fans (EDFs), each capable of generating a maximum thrust of 5 kg at a reference voltage of 50 V. These thrusters are powered by APD 120F3 electronic speed controllers (ESCs), which support up to 12S (50.4 V) input and 120 A continuous current. The entire system is powered by two 6S lithium polymer batteries connected in series, delivering a nominal operating voltage of 50 V when fully charged.

Control of the robot is implemented using Simulink Real-Time, which interfaces through EtherCAT communication with the xPC Target computer, ELMO motor drives, and an STM32F429ZI Nucleo board. This microcontroller is paired with a Hilscher netX board, enabling real-time communication between EtherCAT and non-EtherCAT devices most notably, the ESCs (which require PWM signals) and the motion capture system (communicating via serial interface). Furthermore, the STM32F429ZI communicates with an NVIDIA Jetson Orin Nano, which serves as a high-performance onboard computer for executing computationally intensive tasks such as model predictive control (MPC) and localization. The low-level control loop operates at 500 Hz to ensure precise joint tracking, while the motion capture system provides pose updates at 240 Hz.

\section{Electronics and Firmware}

A structured, multi-level communication framework is implemented within the Hapry robot for precise motion control and state estimation. At the highest level, the motion capture system consists of eight OptiTrack cameras operating at a frequency of 360 Hz, which record infrared reflections from retroreflective markers mounted on the robot’s body. These cameras are connected through Power over Ethernet (PoE+) links, each providing up to 30 W of power per port while simultaneously supporting Gigabit Ethernet data transmission for high-speed, synchronized capture. The motion capture data is transmitted from the OptiTrack host computer to a Linux computer via the NatNet protocol, which enables efficient real-time streaming of six-degree-of-freedom (6-DoF) pose information. This includes both position vectors and quaternion-based orientations, with the overall system achieving a low latency of approximately 2.8 ms.

The Linux machine processes the incoming pose information within the ROS framework and subsequently transmits control commands to an ESP32 microcontroller over Ethernet using the UDP protocol at a rate of 300 Hz. The use of UDP ensures low-latency communication by avoiding the connection overhead inherent to TCP, while a streamlined 8-byte header format is employed to enable efficient and reliable data transfer. The ESP32 interfaces with the STM32 F429ZI microcontroller via the I²C protocol. During initial testing, the system exhibited communication instability attributed to static charge interference. This issue was mitigated by incorporating 5 k$\Omega$ pull-up resistors on both the SDA and SCL lines, thereby ensuring consistent and reliable data transmission.

The STM32 F429ZI microcontroller integrates position data received from the ESP32 with orientation measurements from the onboard IMU, consolidating this information for transmission to the target computer via the EtherCAT network. This architecture enables synchronized control and state feedback across the robot’s six actuated joints. In parallel, the STM32 generates PWM signals for the electronic speed controllers (ESCs) that drive the left and right thrusters, thereby regulating thrust output. Collectively, this establishes the complete control chain—from high-level motion capture input to low-level actuator commands—facilitating coordinated control of both joint dynamics and thrust generation.

An ICM-20948 nine-axis IMU is interfaced with the STM32 via the SPI protocol, delivering orientation data at a frequency of 200 Hz. During early integration, the sensor exhibited electromagnetic interference (EMI) arising from its proximity to the Elmo motor drives; this was effectively mitigated by applying copper tape shielding. The IMU further supports configurable measurement ranges and digital filtering options, allowing performance to be optimized for the robot’s dynamic conditions. In addition, the STM32 communicates with a Hilscher netX netSHIELD board over SPI, which provides EtherCAT connectivity. The netSHIELD module incorporates dual RJ45 ports capable of supporting both line and ring topologies, thereby enabling robust, real-time industrial Ethernet communication.

An ICM-20948 nine-axis IMU is interfaced with the STM32 via the SPI protocol, delivering orientation data at a frequency of 200 Hz. During early integration, the sensor exhibited electromagnetic interference (EMI) arising from its proximity to the Elmo motor drives; this was effectively mitigated by applying copper tape shielding. The IMU further supports configurable measurement ranges and digital filtering options, allowing performance to be optimized for the robot’s dynamic conditions. In addition, the STM32 communicates with a Hilscher netX netSHIELD board over SPI, which provides EtherCAT connectivity. The netSHIELD module incorporates dual RJ45 ports capable of supporting both line and ring topologies, thereby enabling robust, real-time industrial Ethernet communication.

The EtherCAT network links the six Elmo servo amplifiers and the STM32 netX Shield in a daisy-chain configuration, with the chain terminating at a Speedgoat real-time target computer. The Elmo amplifiers natively support advanced motion control features and integrate seamlessly within the EtherCAT framework. The overall control architecture is implemented on a Simulink Real-Time platform, which facilitates rapid prototyping and hardware-in-the-loop (HIL) testing. This setup ensures deterministic execution of complex control algorithms while providing compatibility with a range of industrial communication interfaces, including EtherCAT.

\section{Gantry for constraining frontal dynamics}

\begin{figure}[H]
  \centering
    \includegraphics[width=1\textwidth]{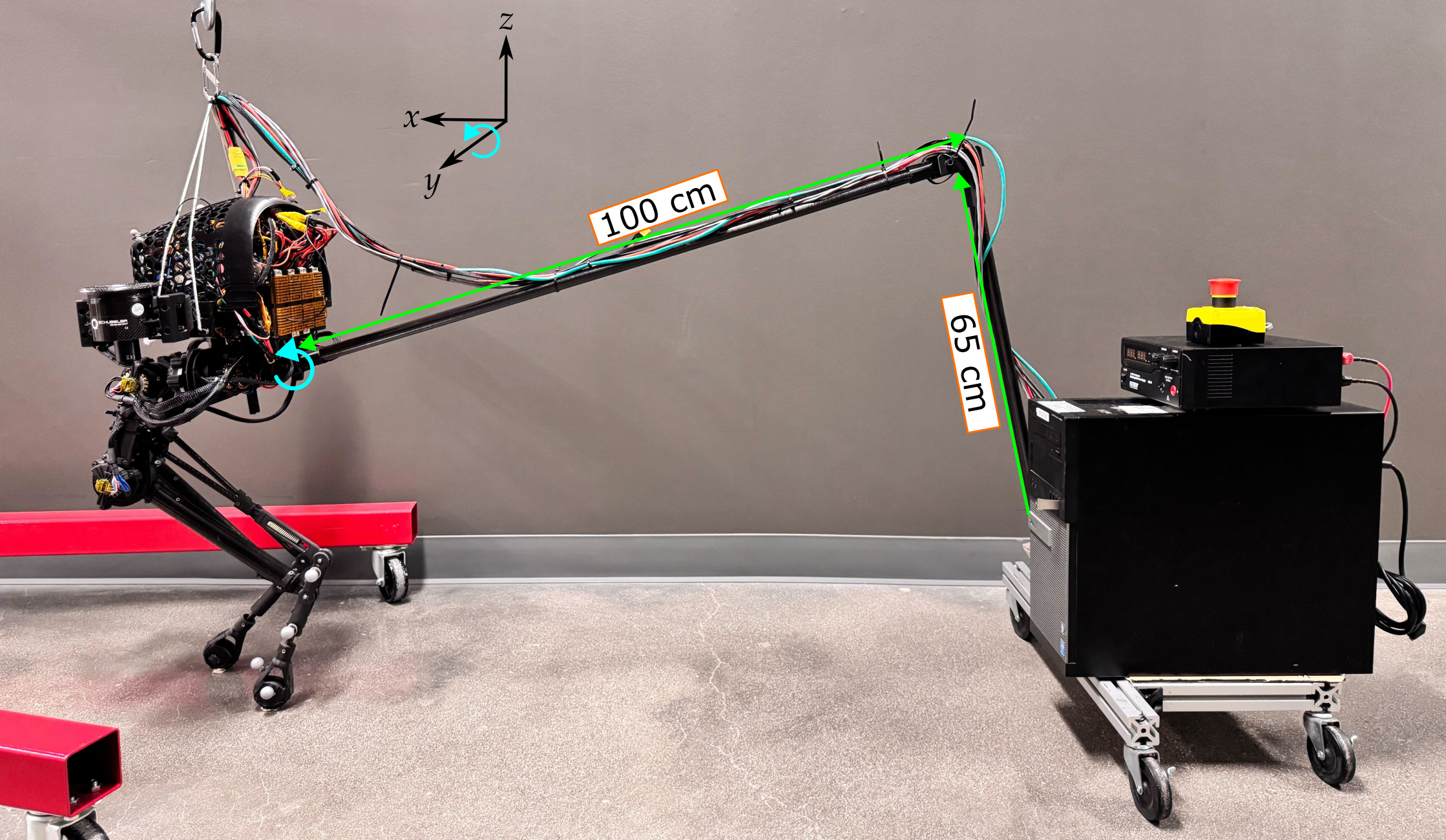}
  \caption{Illustrates Harpy's gantry which contains external electronics, power supply and E-stop}
  \label{fig:harpy-gantry}
\end{figure}

To isolate the sagittal-plane dynamics of the Harpy robot, a lightweight gantry system was developed to constrain motion in the frontal plane. The gantry weighs approximately 300 g and is constructed from carbon fiber, ensuring minimal external disturbance and a negligible contribution to the robot’s overall mass during experiments. In addition to serving as a structural frame for housing essential electrical components, the gantry restricts lateral translation as well as yaw and roll rotations. This configuration enables experiments to focus exclusively on sagittal-plane motion, thereby facilitating a systematic investigation of leg–thruster interactions in locomotion tasks such as trotting and jumping. By eliminating out-of-plane dynamics, the setup allows for a more controlled evaluation of stability, energy distribution, and coordination strategies, which is essential for refining control approaches before extending to full three-dimensional motion.

% Harpy_Mathematical_formulation
\chapter{Modeling}
\label{chap:Harpy-eom}
\section{Energy-based Lagrange Formalism}

This section presents the mathematical formulation for the dynamics of Harpy. The derivation employs the Euler-Lagrange formalism to obtain the equations of motion for a system with 12 DoF, incorporating both joint actuation and thruster.

\subsection{System Description and Modeling Assumptions}

\begin{figure}[h]
  \centering
    \includegraphics[width=0.7\textwidth]{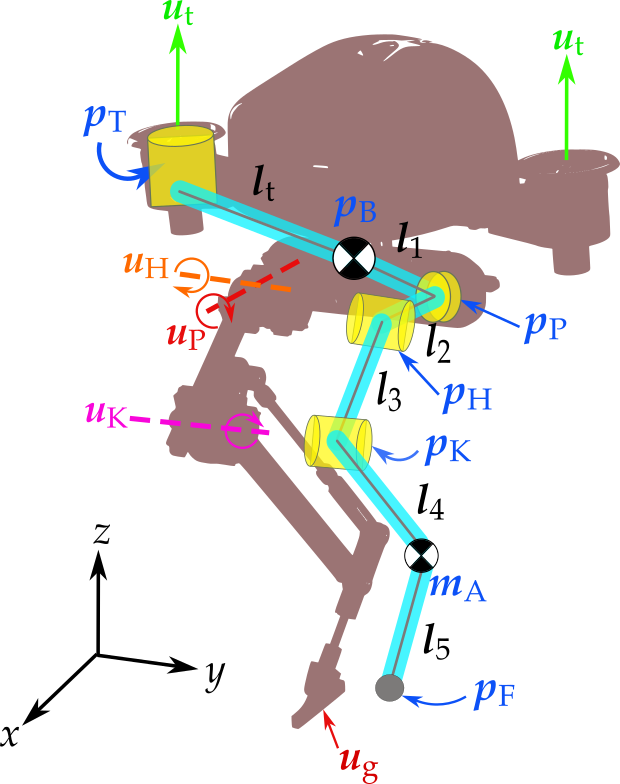}
  \caption{Illustrates Harpy's parameters used for modeling}
  \label{fig:harpy-mdl-params}
\end{figure}

Harpy consists of a central Torso equipped with integrated thrusters and two legs (as shown in \ref{fig:harpy-mdl-params}), each possessing three DoF: hip frontal joint (for abduction \& adduction), hip sagittal joint (forward \& backward), and knee joint (flexion \& extension). The lower legs incorporate parallel linkage mechanisms to maintain foot orientation during locomotion.

To facilitate the dynamics formulation, several simplifying assumptions are adopted. All linkages are considered massless, with mass concentrated at the body center, joint actuators and ankle joint. The lower leg kinematic chain is assumed to have negligible mass, significantly simplifying the system dynamics. Additionally, the pelvis motor mass is incorporated into the body mass for computational efficiency.

\subsection{Kinematic Formulation}

The kinematic description begins with the definition of coordinate frames. The inertial reference frame is denoted as $\{I\}$, while body-fixed frames are established at the body $\{B\}$, hip frontal (pelvis) $\{P\}$, hip sagittal $\{H\}$, knee $\{K\}$, and ankle $\{A\}$ locations. The joint angles are denoted as $\gamma_p$ for hip frontal joint, $\phi_h$ for hip sagittal joint, and $\phi_k$ for knee.

The position vectors of the center of mass for each component are derived through sequential kinematic transformations:
\begin{align}
    \bm{p}_P &= \bm{p}_B + \bm{R}_B \bm{l}_1^B \\
    \bm{p}_H &= \bm{p}_P + \bm{R}_B \bm{R}_x(\gamma_p) \bm{l}_2^P \\
    \bm{p}_K &= \bm{p}_H + \bm{R}_B \bm{R}_x(\gamma_p) \bm{R}_y(\phi_h) \bm{l}_3^H \\
\end{align}
where $\bm{R}_B \in SO(3)$ represents the rotation matrix from the body frame to the inertial frame, $\bm{R}_x(\cdot)$ and $\bm{R}_y(\cdot)$ denote elementary rotation matrices about the $x$ and $y$ axes respectively, and $\bm{l}_i^j$ represents constant length vectors.

The foot and thruster positions are computed as:
\begin{align}
    \bm{p}_F &= \bm{p}_K + \bm{R}_B \bm{R}_x(\gamma_p) \bm{R}_y(\phi_h) \bm{R}_y(\phi_k) \bm{l}_4^K \\
    \bm{p}_T &= \bm{p}_B + \bm{R}_B \bm{l}_t^B
\end{align}
The parallel linkage mechanism in the lower leg yields a kinematic constraint expressed as:
\begin{equation}
    \bm{l}_4^K = \begin{bmatrix}
        -l_{4a} \cos(\phi_k) \\
        0 \\
        -(l_{4b} + l_{4a} \sin(\phi_k))
    \end{bmatrix}
\end{equation}
The angular velocities are found using the following equations:
\begin{align}
    \bm{\omega}_H^B &= \begin{bmatrix} \dot{\gamma}_h & 0 & 0 \end{bmatrix}^{\top} + \bm{\omega}_B^B & \,
    \bm{\omega}_K^H &= \begin{bmatrix} 0 & \dot{\phi}_h & 0 \end{bmatrix}^{\top} + \bm{\omega}_H^H
\end{align}
\subsection{Lagrangian Dynamics Formulation}

The dynamics formulation begins with the construction of energy functions. The total kinetic energy of the system encompasses both translational and rotational components:
\begin{equation}
    \mathcal{K} = \frac{1}{2} \sum_{i \in \mathcal{F}} \left( m_i \bm{\dot{p}}_i^T \bm{\dot{p}}_i + \bm{\omega}_i^{i^T} \bm{I}_i \bm{\omega}_i^i \right)
    \label{eq:KE}
\end{equation}
where $\mathcal{F} = \{B, H_L, K_L, H_R, K_R\}$ represents the set of relevant frames and their associated mass components, with subscripts $L$ and $R$ denoting left and right sides respectively, and $\bm{I}_i$ represents the inertia tensor expressed in the local frame.

The potential energy due to gravitational effects is expressed as:
\begin{equation}
    \mathcal{V} = -\sum_{i \in \mathcal{F}} m_i \bm{p}_i^T \begin{bmatrix} 0 \\ 0 \\ -g \end{bmatrix}
    \label{eq:PE}
\end{equation}
where $g$ represents the gravitational acceleration constant. The Lagrangian of the system is constructed as 
\begin{equation}
    \mathcal{L} = \mathcal{K} - \mathcal{V}.
\end{equation}
To avoid modifying the Lagrangian for rotation matrix which is in $SO(3)$. we use the Euler angle in $XYZ$ formation. Thus the generalized coordinates are as follows: 
\begin{equation}
\bm{q} = [\bm{p}_B^T, \bm \Theta_{B}^{\top}, \gamma_{hL}, \gamma_{hR}, \phi_{hL}, \phi_{hR}, \phi_{kL}, \phi_{kR}]^T
\end{equation}
Euler-Lagrange equations apply:
\begin{equation}
    \frac{d}{dt}\left(\frac{\partial \mathcal{L}}{\partial \dot{\bm{q}}}\right) - \frac{\partial \mathcal{L}}{\partial \bm{q}} = \bm{Q}
    \label{eq:eom}
\end{equation}
\subsection{Equations of Motion in Matrix Form}

The complete equations of motion can be expressed in compact matrix form as:

\begin{equation}
    \bm{M}(\bm{q}) \ddot{\bm{q}} + \bm{h}(\bm{q}, \dot{\bm{q}}) = \bm{B}_j \bm{u}_j + \bm{B}_t \bm{u}_t + \bm{B}_g \bm{u}_g
\end{equation}

where $\ddot{\bm{q}} = [\ddot{\bm{p}}_B^T, \ddot{\bm \Theta_{B}}^{\top}, \ddot{\gamma}_{hL}, \ddot{\gamma}_{hR}, \ddot{\phi}_{hL}, \ddot{\phi}_{hR}, \ddot{\phi}_{kL}, \ddot{\phi}_{kR}]^T$ represents the system accelerations, $\bm{M}(\bm{q})$ is the configuration-dependent mass matrix, and $\bm{h}(\bm{q}, \dot{\bm{q}})$ contains Coriolis, centrifugal, and gravitational terms. The inputs consist of joint actuation torques $\bm{u}_j$, thruster forces $\bm{u}_t$, and ground reaction forces $\bm{u}_g$.The thruster force vector is composed as $\bm{u}_t = [\bm{u}_{tL}^T, \bm{u}_{tR}^T]^T$, representing left and right thruster contributions.

The input matrices are constructed through the principle of virtual work. The joint actuation matrix takes the form:

\begin{equation}
    \bm{B}_j = \begin{bmatrix}
        \bm{0}_{6 \times 6} \\
        \bm{I}_{6 \times 6}
    \end{bmatrix}
\end{equation}

allowing direct actuation of the joint angles. The thruster and ground force input matrices are derived from the Jacobian relationships:
\begin{equation}
\begin{aligned}
    B_t = \begin{bmatrix}
        \begin{pmatrix}
        \partial \dot{\bm p}_{T_L} / \partial \bm v \\
        \partial \dot{\bm p}_{T_R} / \partial \bm v
        \end{pmatrix}^\top
        \\
        0_{2 \times 6}
    \end{bmatrix}, \quad
    B_g = \begin{bmatrix}
        \begin{pmatrix}
        \partial \dot{\bm p}_{F_L} / \partial \bm v \\
        \partial \dot{\bm p}_{F_R} / \partial \bm v
        \end{pmatrix}^\top
        \\
        0_{2 \times 6}
    \end{bmatrix}.
\end{aligned}
\label{eq:generalized_forces}
\end{equation}
where $\bm{v} = [\bm{\omega}_B^{B^T}, \dot{\bm{q}}^T]^T$ represents the generalized velocity coordinates.

\subsection{Ground Contact Modeling}

The interaction between the robot's feet and the ground surface is modeled using a unilateral compliant contact framework. The ground reaction forces are computed as:

\begin{equation}
    \bm{u}_g = \begin{cases}
        k_g \delta_n \bm{n} - b_g \dot{\delta}_n \bm{n} + \bm{f}_f & \text{if } \delta_n > 0 \\
        \bm{0} & \text{if } \delta_n \leq 0
    \end{cases}
\end{equation}

where $\delta_n$ represents the normal penetration depth, $k_g$ and $b_g$ denote the ground stiffness and damping coefficients respectively, $\bm{n}$ is the contact normal vector, and $\bm{f}_f$ represents the friction force computed according to an appropriate friction model.

The presented equations of motion provide a complete dynamic model for Harpy, incorporating both traditional joint actuation and thruster forces. This formulation provides a modular structure suitable for simulation, control design, and trajectory optimization for hybrid locomotion combining walking and jumping maneuvering capabilities.

In experimental validation and certain operational scenarios, the robot is attached to a gantry system that restricts its motion to a vertical plane. This section derives the modified equations of motion accounting for these kinematic constraints imposed by the gantry mechanism.

\section{Gantry Constraint Formulation}
For Trotting and Jumping experiments to study the sagittal dynamics, we use gantry shown in Fig.~\ref{fig:harpy-gantry} to enforce planar constrained on hardware.The gantry system enforces planar motion by constraining the robot's body position and orientation. Specifically, the gantry eliminates lateral translation along the $y$-axis and prevents rotation about the roll and yaw axes. To apply similar constraint on the EoM derived in the above section~\eqref{eq:eom}. We start by explicitly writing the constrained enforced.

\begin{equation}
    P_{B,y} = 0, \quad \theta_{\text{roll}} = 0, \quad \theta_{\text{yaw}} = 0
\end{equation}

where $P_{B,y}$ denotes the $y$-component of the body position, and $\theta_{\text{roll}}$ and $\theta_{\text{yaw}}$ represent the roll and yaw angles respectively. These holonomic constraints can be expressed in terms of the configuration variables through the constraint function:

\begin{equation}
    \bm{f}(\bm{q}) = \bm{0}_{3 \times 1}
\end{equation}

The velocity-level constraints are obtained by differentiating the position constraints with respect to time:

\begin{equation}
    \frac{\partial \bm{f}}{\partial \bm{q}} \dot{\bm{q}} = \bm{J}_c \dot{\bm{q}} = \bm{0}_{3 \times 1}
\end{equation}

where $\bm{J}_c$ represents the constraint Jacobian matrix. Similarly, the acceleration-level constraints are derived through further differentiation:

\begin{equation}
    \frac{\partial^2 \bm{f}}{\partial t^2} = \bm{J}_c \ddot{\bm{q}} + \dot{\bm{J}}_c \dot{\bm{q}} = \bm{0}_{3 \times 1}
    \label{eq:acc-const}
\end{equation}

\subsection{Incorporation of Constraint Forces}

The presence of the gantry introduces constraint forces that maintain the prescribed kinematic restrictions. Using the method of Lagrange multipliers, the equations of motion are augmented to include these constraint forces:

\begin{equation}
    \bm{M} \ddot{\bm{q}} + \bm{h}(\bm{q}, \dot{\bm{q}}) = \bm{B} \, \bm{u} + \bm{J}_c^T \bm{\lambda}
    \label{eq:aug-eom}
\end{equation}

where $\bm{B} = [\bm B_j^{\top},\, \bm B_t^{\top},\, \bm B_g^{\top}]$, $\bm{u} = [\bm u_j^{\top},\, \bm u_t^{\top},\, \bm u_g^{\top}]^{\top},$ represent the combined input and input matrices. $\bm{\lambda} = [\lambda_1, \lambda_2, \lambda_3]^T$ represents the vector of Lagrange multipliers corresponding to the three constraints. These multipliers physically represent the constraint forces and moments applied by the gantry to enforce the planar motion.

\subsection{Constrained System Dynamics}

Combining the augmented EoM~\eqref{eq:aug-eom} with the acceleration-level~\eqref{eq:acc-const} constraints yields the following system of differential-algebraic equations (DAEs):

\begin{equation}
    \begin{bmatrix}
        \bm{M} & \bm{J}_c^T \\
        \bm{J}_c & \bm{0}
    \end{bmatrix}
    \begin{bmatrix}
        \ddot{\bm{q}} \\
        \bm{\lambda}
    \end{bmatrix}
    =
    \begin{bmatrix}
        \bm{B} \bm{u} - \bm{h} \\
        -\dot{\bm{J}}_c \dot{\bm{q}}
    \end{bmatrix}
\end{equation}

This system simultaneously solves for the accelerations $\ddot{\bm{q}}$ and the constraint forces $\bm{\lambda}$. The matrix on the left-hand side is symmetric and, under appropriate conditions, invertible, allowing for unique determination of both the motion and constraint forces.

The gantry-constrained formulation enables controlled experimental validation of the robot's dynamics while maintaining the fidelity of the mathematical model. This approach facilitates the transition from constrained laboratory testing to free-flight operation by providing consistent dynamic models for both scenarios.

% Results_discussion
\chapter{Result and Discussion}
The experiments were conducted using a system set-up that comprising the OptiTrack motion capture system, Harpy platform with the gantry set-up and a target PC. The OptiTrack system provided ground-truth body position and orientation, as well as foot marker trajectories, while onboard sensors delivered joint positions and velocities from encoders and joint currents from the amplifiers. The target PC recorded the desired kinematic trajectories generated by the Raibert controller, as well as the commanded joint positions and velocities. Additional logged signals included thruster PWM inputs (later mapped to thrust forces via calibration), estimated world-frame and body-frame foot end positions and  state-machine phases. All measurements were timestamped and synchronized across systems, enabling direct comparison between desired trajectories, EKF-based state estimates, and ground-truth OptiTrack data. Orientation estimates were obtained through an EKF, which fused data from an Adafruit ICM-20948 nine-axis IMU sampled at 500 Hz with OptiTrack rigid body motion capture measurements collected at 300 Hz.
\section{Trotting Experimental results}
This section presents the experimental findings from our analysis of harpy’s trotting dynamics across 11 gait cycles or 3.5 seconds of trotting. From this experiment quantified key parameters including body position, body orientation, foot placement, joint position, joint velocity, UDoF. The following subsections detail our observations from the collected data.
\begin{figure}[H]
  \centering
    \includegraphics[width=1\textwidth]{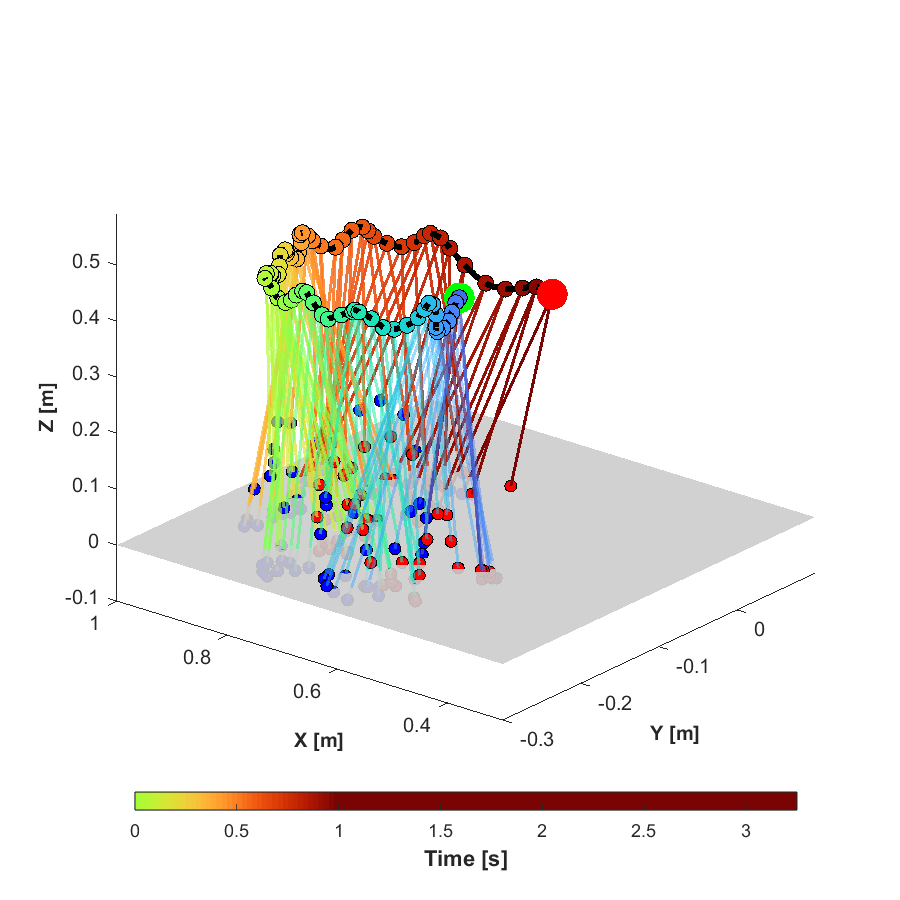}
  \caption{Illustrates the hardware trajectory overlayed on the stick diagram. Color gradient shows the time ranging from $0$-$3.5$ sec.}
  \label{fig:Trot-stick-model}
\end{figure}
The Figure\ref{fig:Trot-stick-model} illustrates the 3D stick diagram representation of Harpy’s trotting experiment under gantry-constrained conditions, where the temporal progression of the COM and leg trajectories is visualized with time-based color mapping. The vertical body motion ranges from 0.4 m to 1 m, while the relative displacement with respect to the stance leg is confined between –0.2 m and 0.2 m, indicating stable sagittal-plane dynamics. Only minor lateral deviations are observed in the Y-direction, which can be attributed to small slackness in the gantry and the fact that ideal rigid-body conditions are not fully realized. Nevertheless, these deviations remain minimal, highlighting the gantry’s effectiveness in constraining lateral dynamics. Small COM oscillations are present, primarily caused by ankle spring compliance, which introduces slight elastic dynamics during trotting. Event timing, including the swing and stance phases, remains consistent across steps with minimal variance, and the robot maintains symmetric step lengths, reinforcing gait regularity and stability. In addition, some foot sinkage into the ground is visible, which can be attributed to structural compliance and the simplified rigid-body kinematic model used in evaluation. Overall, the stick diagram demonstrates that the gantry successfully isolates sagittal-plane motion while permitting natural trotting sequences, providing stable locomotion dynamics with minimal disturbance in other directions and validating it as a reliable testbed for locomotion studies.
\begin{figure}[H]
  \centering
    \includegraphics[width=1\textwidth]{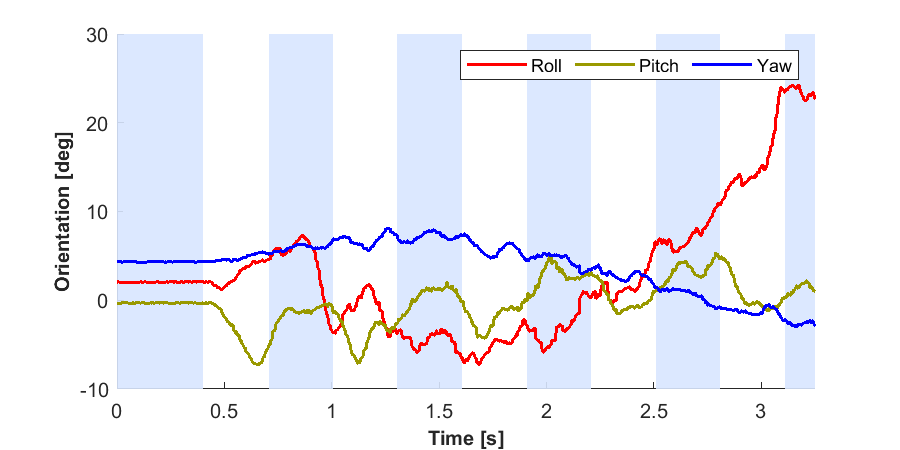}
  \caption{Illustrates orientation of robot during trotting}
  \label{fig:Trot-orient}
\end{figure}
This Figure\ref{fig:Trot-orient} presents the evaluation of body orientation during the trotting experiment, with roll, pitch, and yaw trajectories plotted over time. The blue shaded regions correspond to the swing phases of the gait cycle. The pitch remains bounded between –5° and +5°, which demonstrates stable forward motion without excessive tilting. Similarly, both roll and yaw stay within –10° to +10° for approximately 2.5 seconds, indicating good lateral balance and heading stability during this period.
Orientation data is obtained through an EKF framework, which fuses measurements from an Adafruit ICM-20948 nine-axis IMU operating at 500 Hz with OptiTrack motion capture data sampled at 300 Hz. Allan variance analysis was performed on the IMU signals to identify noise parameters, ensuring accurate tuning of the estimator. By leveraging EKF fusion, the orientation estimates appear smooth and robust, minimizing sensor noise while capturing dynamic body motion. For validation, the actual orientation from OptiTrack was compared against the estimated orientation from the EKF, and the error between them was computed and analyzed.
Overall, these results confirm that the robot maintains bounded orientation while trotting, and the EKF-based fusion provides reliable state estimates that are essential for stable and repeatable locomotion.
\begin{figure}[H]
  \centering
    \includegraphics[width=1\textwidth]{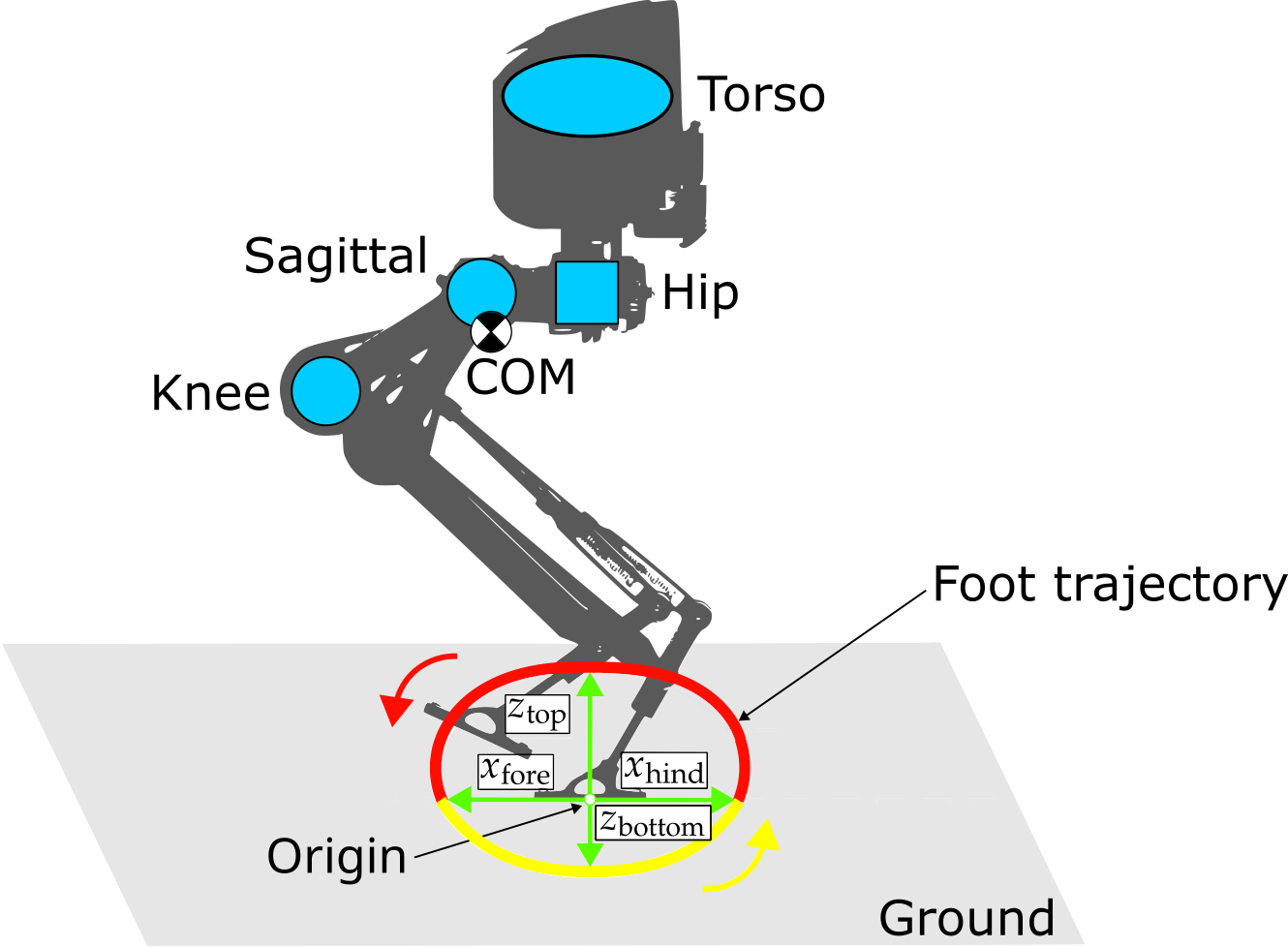}
  \caption{Illustrates the method used for characterizing tracking performance}
  \label{fig:fig_foot_trot}
\end{figure}
To systematically analyze foot placement during trotting, the foot trajectory was parameterized using four primary variables: \(z_{\text{top}}\) and \(z_{\text{bottom}}\), which represent the maximum and minimum vertical excursions of the trajectory, and \(x_{\text{fore}}\) and \(x_{\text{hind}}\), which capture the forward and backward reach of the step in the sagittal plane. 
\begin{equation}
\mathbf{P}_{BF} = FK(q_j) \quad \quad 
\mathbf{P}_{NB} = \begin{bmatrix} 0 \\ -0.05 \\ -0.5 \end{bmatrix}
\end{equation}
\begin{equation}
\mathbf{P}_{FN} = \mathbf{P}_{BF} - \mathbf{P}_{NB}
\end{equation}
As illustrated in Fig.\ref{fig:fig_foot_trot}, the formulation is based on three vectors that together describe the foot’s relative motion. The vector \(P_{\text{BF}}\) extends from the torso center to the actual foot location, shown in red to indicate the swing foot position. The vector \(P_{\text{NB}}\) extends from the torso center to the initial foot location in the body frame, shown in yellow to represent the stance foot. The displacement vector \(P_{\text{FN}}\) is then defined as the difference between these two, capturing the deviation of the actual foot position from the initial reference location. This decomposition provides a clear representation of how each step evolves relative to both the robot body and the initial placement of the foot. By adopting this formulation, the trajectory of the foot can be precisely expressed in terms of meaningful parameters. Positive displacements along the z-axis correspond to \(z_{\text{top}}\), marking the apex of the foot trajectory, while negative displacements correspond to \(z_{\text{bottom}}\), representing the lowest point of the swing. Similarly, displacements along the x-axis are associated with \(x_{\text{fore}}\) (forward extension) and \(x_{\text{hind}}\) (rearward extension), which together define the step length.  This parameterization allows systematic control of step trajectories and provides a framework to study their effects on gait repeatability, stability, and locomotion performance.
\begin{figure}[H]
  \centering
    \includegraphics[width=1\textwidth]{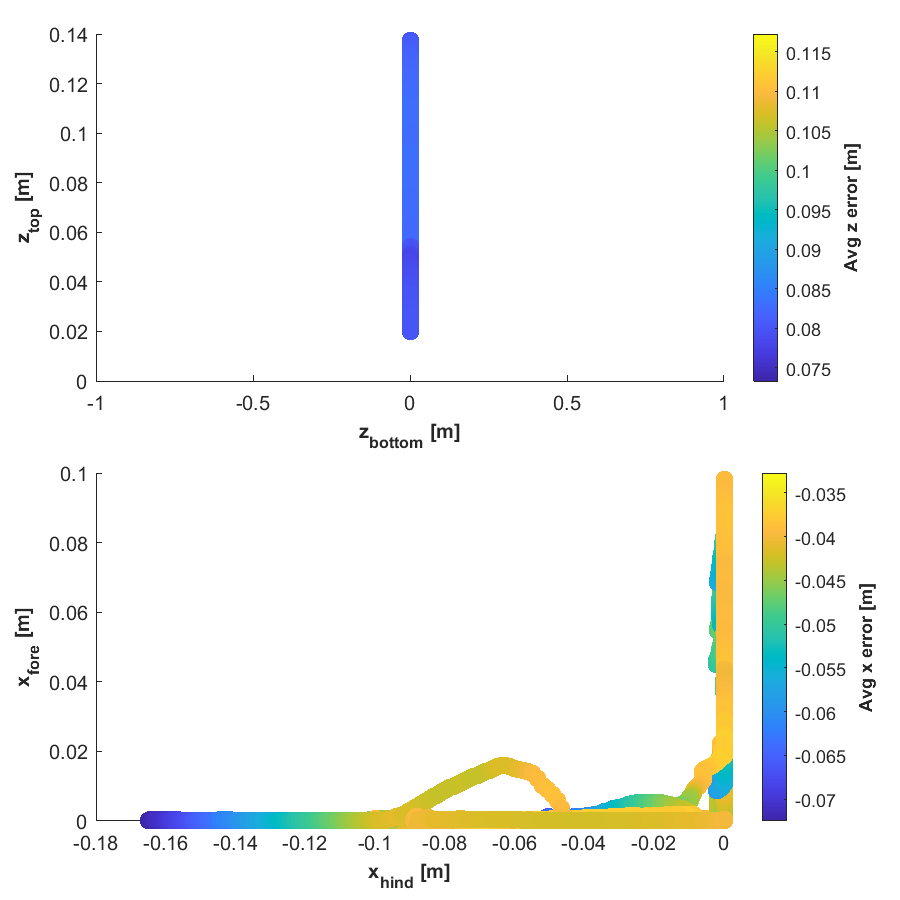}
  \caption{Illustrates the tracking error (shown as color gradient) in x and z component of foot position}
  \label{fig:Trot-kin-limit}
\end{figure}
The stability of trotting motion was further evaluated through the analysis of limit cycles in the underactuated degrees of freedom (UDoFs), specifically the body’s sagittal and vertical dynamics. Figure~\ref{fig:Trot-kin-limit} shows phase plots of body velocity versus body position, where continuous trajectories form bounded loops. The presence of smooth and continuous loops indicates that the system naturally exploits passive dynamics rather than relying on aggressive corrective control. Multiple concentric loops are observed in the vertical (z) direction, suggesting slight step-to-step variability while still maintaining an overall repeatable pattern. The boundedness of underactuated states such as the COM and body altitude reflects stable oscillations that remain confined rather than diverging over time. Furthermore, the elliptical shape of the loops confirms the existence of closed orbits, which are characteristic of stable locomotion in limit-cycle systems. Overall, the UDoF limit cycle analysis demonstrates that despite underactuation, the robot maintains bounded and repeatable dynamics during trotting, validating the stability of its gait. These results also reinforce the role of the lightweight gantry in isolating sagittal-plane motion, enabling clear observation of bounded limit cycles and providing confidence in the reliability of the experimental setup.
\begin{figure}[H]
  \centering
    \includegraphics[width=1\textwidth]{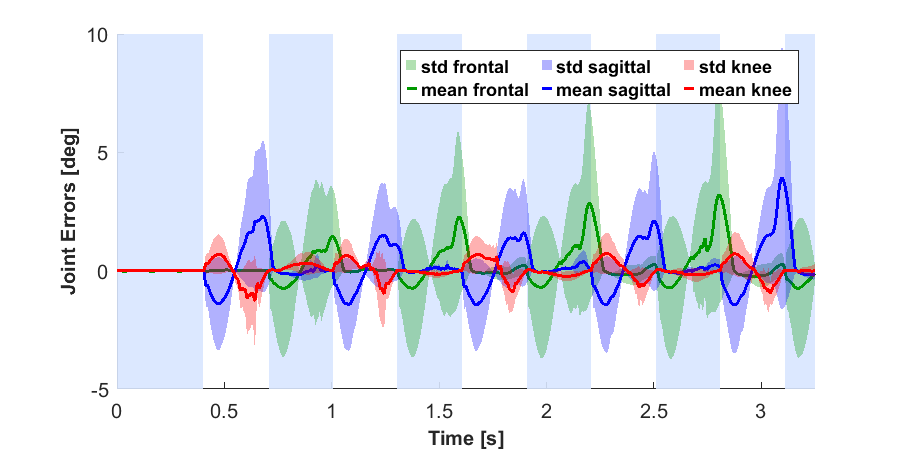}
  \caption{Illustrates the mean and std of joint error for left foot}
  \label{fig:Trot-joint-error}
\end{figure}
To assess the tracking performance of individual joints during trotting, joint angle errors were computed for the frontal, sagittal, and knee joints. These errors were calculated as the difference between the desired and actual joint positions over time, where desired joint positions were generated via inverse kinematics and the actual positions were obtained from joint encoders. For each joint pair (left and right), the values were averaged to obtain a representative joint angle for that degree of freedom. The mean and standard deviation of the errors were then evaluated across the full time span to analyze joint tracking consistency. The results are shown in Fig.~\ref{fig:Trot-joint-error}, with shaded regions representing one standard deviation around the mean error for each joint group.
The frontal joint error ranges from approximately -4 to +8 degrees, while the sagittal joint error ranges from -3.5 to +9.4 degrees. Both of these joints exhibit increasing variability over time, as indicated by the widening of their shaded error bands. In contrast, the knee joint shows a much narrower error range, approximately -3 to +1.5 degrees, and maintains a relatively constant standard deviation throughout the gait cycle. This consistency in the knee joint suggests more reliable impact absorption and a better ability to handle repetitive loading during each stance phase.
Overall, the analysis highlights a clear distinction between the frontal/sagittal joints, which show gradual divergence, and the knee joint, which maintains stable behavior across multiple steps. This reinforces the role of the knee joint in ensuring shock absorption and contributing to consistent locomotion under repeated contact dynamics.
\begin{figure}[H]
  \centering
    \includegraphics[width=1\textwidth]{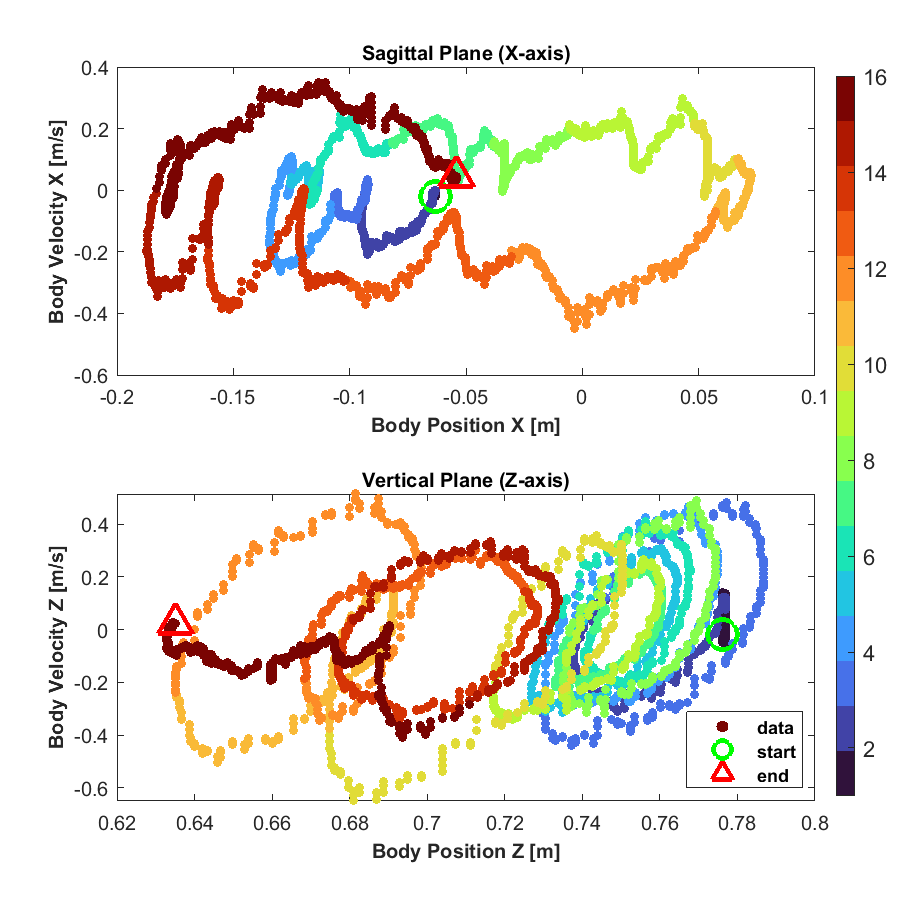}
  \caption{Illustrates phase portrait for Udof component i.e. X and Z of body}
  \label{fig:Trot-body-limit}
\end{figure}
To evaluate the dynamic behavior of the robot's UDoFs during trotting, phase plots were constructed in the sagittal (x) and vertical (z) planes by plotting body velocity against body position. These plots, shown in Fig.~\ref{fig:Trot-body-limit}, reveal the presence of closed-loop trajectories or limit cycles that describe the evolution of the robot's motion over repeated steps. The position and velocity data were obtained from OptiTrack measurement. 
The resulting trajectories are smooth and continuous, indicating that the system leverages its passive dynamics rather than relying on high-gain control. In the vertical plane, multiple concentric loops are observed along the z-axis, representing minor step-to-step variations. Crucially, these oscillations remain bounded throughout the duration of the experiment, with no signs of divergence. This boundedness of underactuated states confirms that the robot remains dynamically stable during trotting. The elliptical shape of the loops further supports the interpretation of a stable limit cycle behavior, where the gait converges to a repeatable and dynamically consistent trajectory over time.
Overall, the presence of such bounded and repeatable limit cycles provides strong evidence that the system achieves stable trotting behavior despite under actuation.

\begin{figure}[H]
  \centering
    \includegraphics[width=1\textwidth]{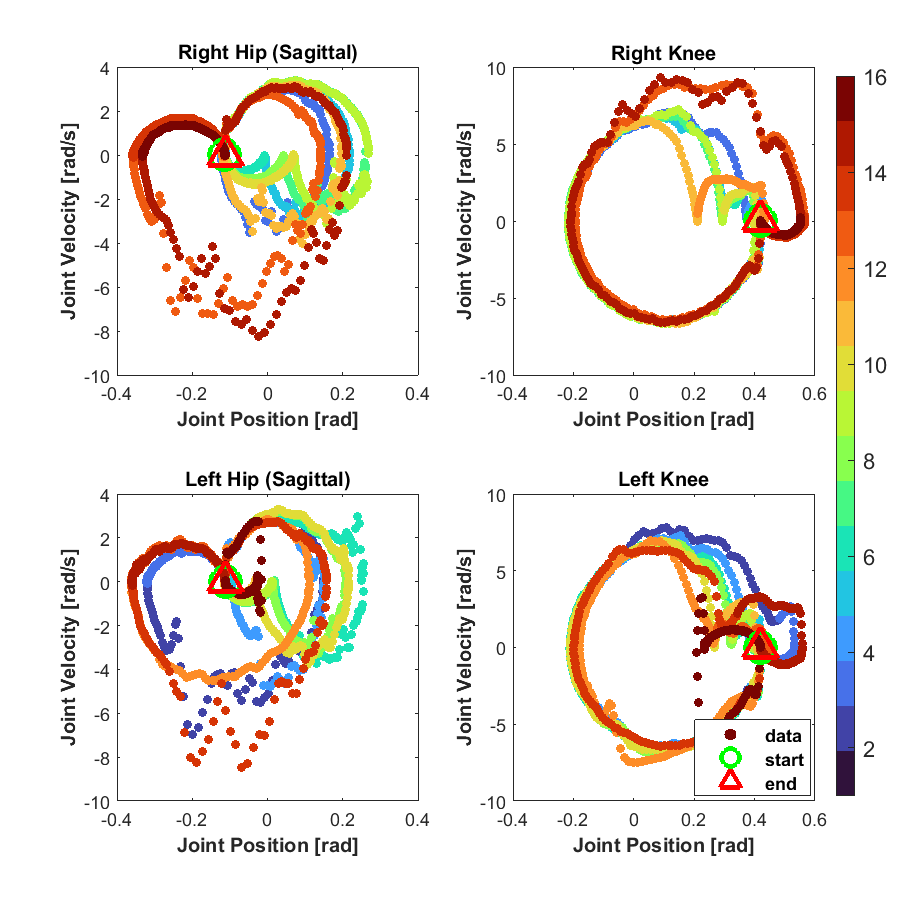}
  \caption{Illustrates phase portrait for actuated DoF i.e., right leg sagittal joint (top left), left leg sagittal joint (bottom left), right leg knee joint (top right), left leg knee joint (bottom right). }
  \label{fig:Trot-joint-limit}
\end{figure}
To further analyze the repeatability and coordination of leg movement during trotting, joint angle limit cycles were constructed for the hip and knee joints in both left and right legs. These phase plots, shown in Fig.~\ref{fig:Trot-joint-limit}, represent joint velocity plotted against joint position over time and reveal consistent closed-loop patterns across multiple gait cycles. The presence of these closed limit cycles is a indicator of stable and repeatable locomotion, indicating that the robot's leg motion converges to a consistent cyclic behavior despite environmental perturbations and model imperfections.
In the sagittal plane of the hip joints, a figure-8 shaped trajectory is observed for both the left and right legs. This pattern is particularly informative, as it highlights the alternation between the swing and stance phases within each gait cycle. During the swing phase, the hip joint undergoes a larger range of motion with increased angular velocity, while the stance phase corresponds to the tighter, more compact loop near the zero-velocity crossing. This smooth transition is repeated across steps and demonstrates effective control of Hip actuation in synchrony with whole-body motion.
The knee joint plots exhibit a distinct dual-cluster structure, with one large loop and a smaller compact cluster. The larger outer loop corresponds to the swing phase, during which the knee extends and flexes dynamically to provide foot clearance and prepare for landing. In contrast, the smaller internal cluster corresponds to the stance phase, where the knee joint remains relatively stationary and stiff to support body weight and absorb impact forces. Additionally, across all joints, the magnitude of joint velocities remains relatively low compared to the mechanical limits of the actuators. This suggests that the system is operating in a dynamically favorable regime cooperating with natural dynamics rather than fighting them. Such behavior is characteristic of energy-efficient locomotion and reflects a control strategy that leverages passive dynamics to minimize unnecessary actuation.
Together, these joint-space limit cycles validate that the robot exhibits stable, symmetric, and energetically favorable leg trajectories during trotting. The repeatability and structure of these loops provide strong evidence that the joint-level control contributes to the overall dynamic stability of the gait.
\section{Jumping Experimental results}
\begin{figure}[H]
  \centering
    \includegraphics[width=1\textwidth]{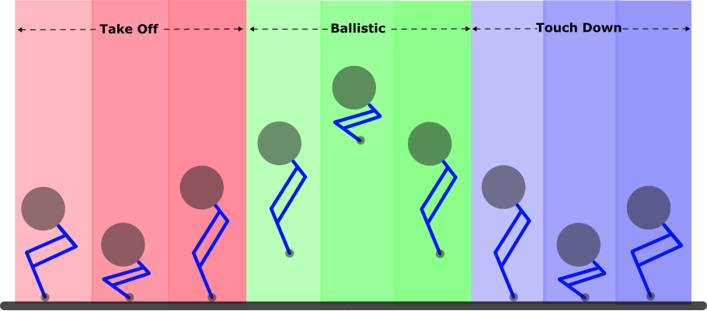}
  \caption{Illustrates the hardware jumping motion data under consideration with different phases represented by color.}
  \label{fig:Jump-mot-ink}
\end{figure}
This section presents the experimental findings from our analysis of harpy's jumping dynamics across $4$ different experiments. We systematically examined the hybrid locomotion characteristics across three distinct phases of the jump cycle: take-off (ground contact with thruster ignition), ballistic (aerial phase with thruster contribution), and touch-down (landing and stabilization), as shown in Fig.~\ref{fig:Jump-mot-ink}. The integration of the thruster force introduced a unique control challenge as compared to conventional robot. The thruster-leg coordination plays a vital role in stable jumping maneuver. Through comprehensive experimental trials, we quantified key parameters including joint torques, thruster forces, attitude stability, center of mass dynamics and individual system's (Leg \& thruster) energy throughout each phase. The following subsections detail our observations from the collected data, with particular emphasis on the coupling effects between thruster activation and bipedal mechanics. 
\begin{figure}[H]
  \centering
    \includegraphics[width=1\textwidth]{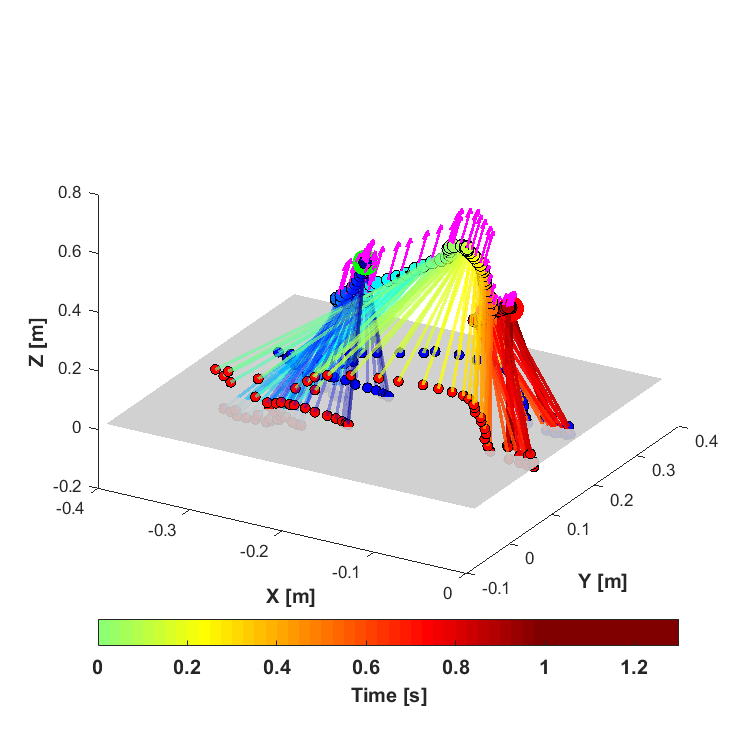}
  \caption{Illustrates the hardware trajectory overlayed on the stick diagram. Color gradient shows the time ranging from $0$-$1.2$ sec.}
  \label{fig:Jump-stick-diag}
\end{figure}
Figure\ref{fig:Jump-stick-diag} illustrates the evidence of stability of underactuated degrees of freedom during thruster-assisted jumping. Analysis of the stick diagram reveals the COM trajectory, which exhibits periodic vertical motion accompanied by alternating foot support. Such periodicity is a strong indicator of rhythmic and repeatable dynamics, reinforcing the notion of bounded motion within the underactuated coordinates. The trajectory further highlights distinct lift-off and landing phases, with the COM consistently achieving its maximum height of 0.2 meters at mid-jump. This observation confirms the effective contribution of thrusters in augmenting vertical propulsion and enabling jump amplitudes higher than would be possible through leg actuation alone. In addition, the symmetry observed between the left and right leg trajectories reflects well-coordinated foot placement, which plays a critical role in maintaining stability during both take-off and landing. Importantly, this symmetry persists even in the presence of external thrust inputs, underscoring the robustness of the system’s coordination strategy. Taken together, these findings demonstrate that Harpy is capable of executing dynamically stable, repeatable, and well-bounded jumping motions under thruster assistance, thereby validating the control framework for maintaining stability under underactuated conditions.
\begin{figure}[H]
  \centering
    \includegraphics[width=1\textwidth]{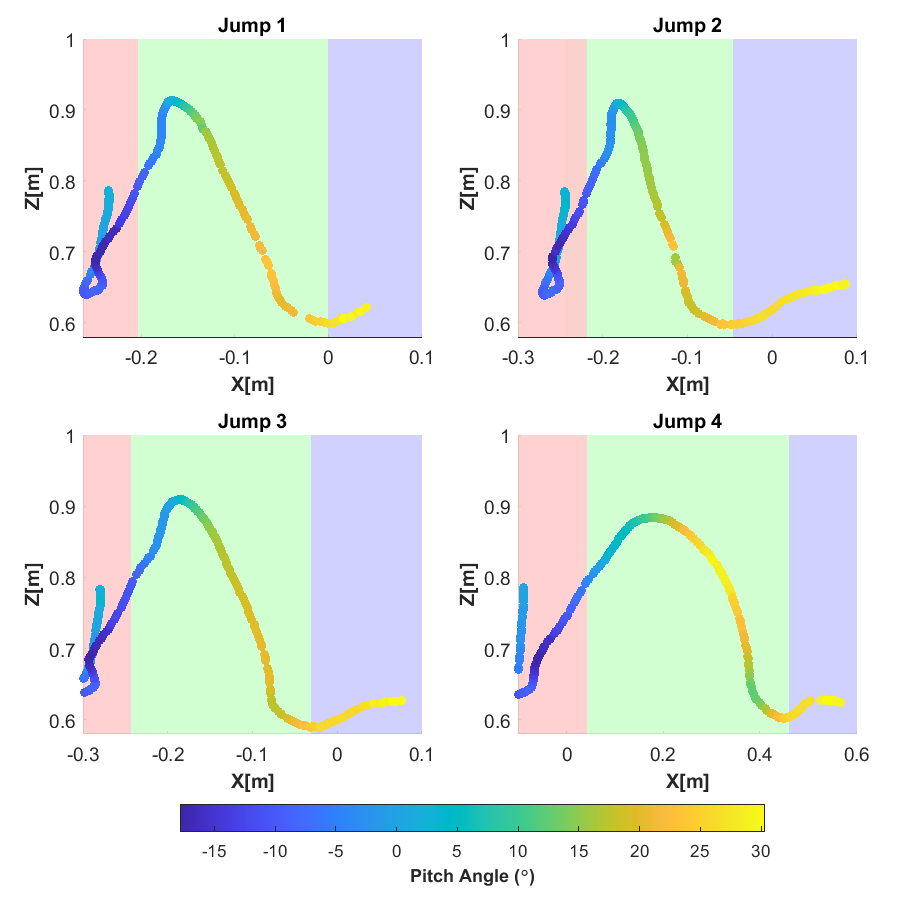}
  \caption{Illustrates the COM trajectory across $4$ experiments. Color gradient indicates the body pitch.}
  \label{fig:Jump-traj}
\end{figure}
Figure\ref{fig:Jump-traj} presents the COM trajectories for $4$ experimental jumping trials of Harpy, with body pitch angle represented through the color gradient. Analysis of these trajectories reveals several key performance characteristics that highlight both the capabilities and challenges of the hybrid locomotion system. All four jumps achieve remarkably consistent peak heights between $0.91-0.92$ m from the initial standing position of approximately $0.65$ m, resulting in a net vertical displacement of $0.26 - 0.27$ m, with maximum altitude reached during the early-to-mid ballistic phase (green region). This consistency indicates good repeatability of the thruster control system and effective coordination between initial leg push-off and thruster activation. However, horizontal displacement patterns show significant variation between trials, ranging from minimal forward drift of approximately $0.1$ m in Jump $1$ to nearly $0.6$ m in Jump $4$. This variation suggests challenges in maintaining consistent thrust vectoring or potentially indicates different experimental conditions between trials. The color-coded pitch angle data reveals distinct dynamics throughout each jump phase: during take-off (red region), the robot maintains a slight forward pitch $(0 \,\text{to}\, -5 \deg)$ that facilitates forward momentum generation; the ballistic phase shows significant pitch variation ranging from $-15 \deg$ (backward tilt) to $+30 \deg$ (forward tilt), with maximum forward pitch typically occurring at the trajectory apex; and touch-down phases generally show the robot returning to near-neutral pitch angles around $ +5\,to\,-5 \deg$. The thruster are mounted onto the torso closer to the COM as shown in Fig.~\ref{fig:harpy-hardware}, due to this thruster cannot produce counter pitching moment. Thus, during the ballistic phase, robot cancels out the forward pitch $-10 \deg$ consistently across multiple experiments by generating the angular momentum from leg swing. This further highlight the thruster vectoring capability of harpy using posture manipulation.
\begin{figure}[H]
  \centering
    \includegraphics[width=1\textwidth]{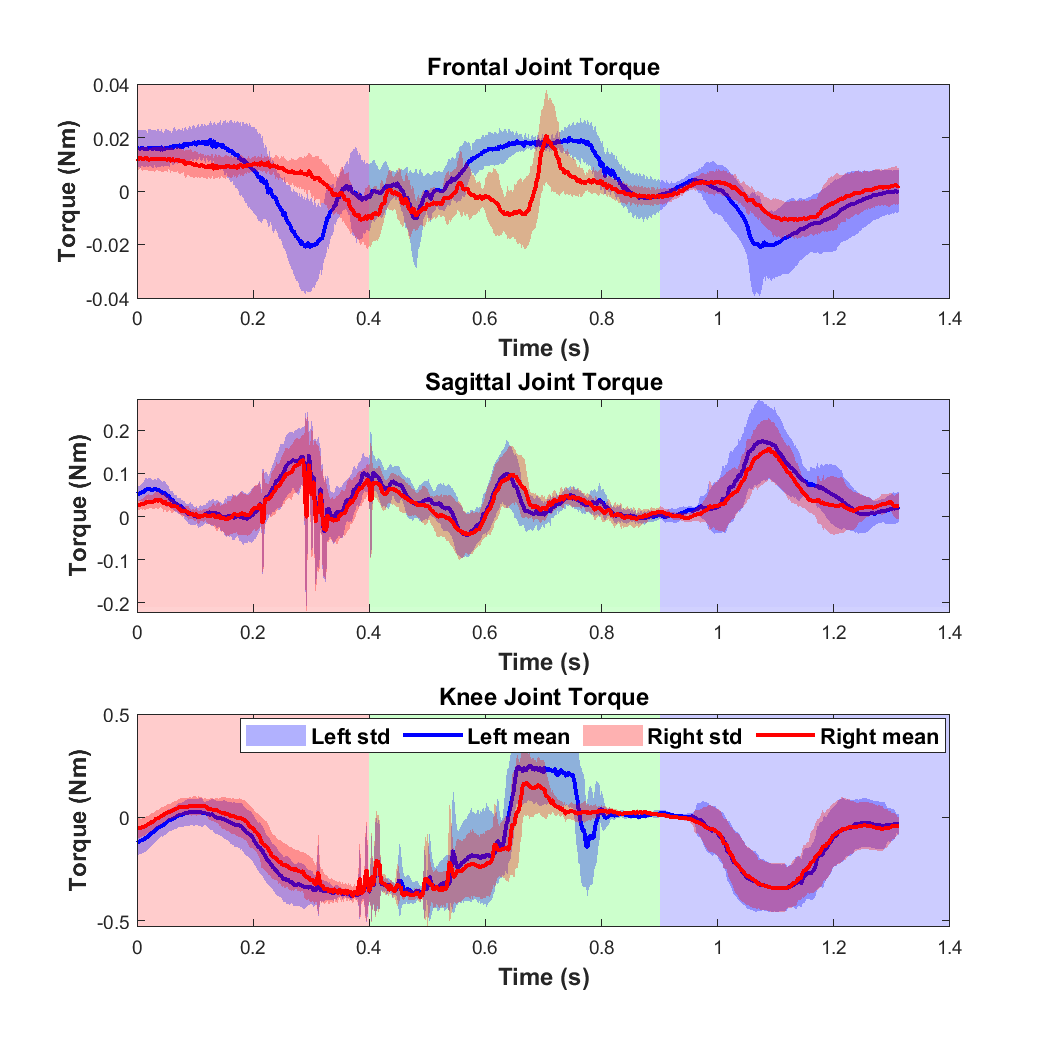}
  \caption{Shows the mean and standard deviation for frontal, sagittal and knee joint for both left (Blue) and right foot (Blue).}
  \label{fig:Jump-torque}
\end{figure}
Figure~\ref{fig:Jump-torque} illustrates the temporal evolution of joint torques for the frontal, sagittal, and knee joints during the thruster-assisted jumping motion, with mean values and standard deviations shown for both left and right legs across multiple trials. The torque profiles reveal distinct patterns that correspond to the three jump phases and highlight the complex biomechanical demands of hybrid locomotion. During the take-off phase $(0-0.4s, \text{red region})$, the knee joints generate the highest torques, reaching peak values of approximately $0.3$ Nm, which is expected as they provide the primary propulsive force for vertical acceleration. The frontal plane torques remain relatively small throughout all phases $(±0.04 \text{Nm})$, indicating minimal lateral stabilization requirements due to gantry applying frontal constraint. The sagittal joint torques show interesting bi-phasic patterns, with initial negative torques $(-0.15 \text{Nm})$ during early take-off transitioning to positive values $(0.1 \text{Nm})$ during late take-off, likely reflecting the forward pitch adjustments needed to coordinate with thruster activation. During the ballistic phase $(0.4-1.0s, \text{green region})$, all joint torques exhibit high-frequency oscillations with reduced magnitudes, as the legs are unloaded but must maintain configuration for landing preparation. The knee torques show particularly interesting behavior during this phase, with intermittent spikes suggesting active position adjustments to counter the initial angular momentum during take-off phase. The touch-down phase $(1.0-1.4s, \text{blue region})$ presents the most demanding torque requirements, with knee joints experiencing rapid torque reversals from $0.3$ Nm to $-0.4$ Nm as they absorb landing impact and stabilize the robot. The sagittal joints similarly show significant torque generation (up to 0.2 Nm) during landing, indicating their crucial role in maintaining anterior-posterior balance. The close correspondence between left and right leg torques throughout the motion demonstrates good bilateral symmetry in the control strategy, though slight variations in the standard deviation bands suggest some trial-to-trial variability in load distribution. These torque profiles collectively indicate that while thrusters assist with vertical propulsion, the leg joints remain critical for jump initiation, aerial configuration control, and landing stabilization in this hybrid locomotion system.
\begin{figure}[H]
  \centering
    \includegraphics[width=1\textwidth]{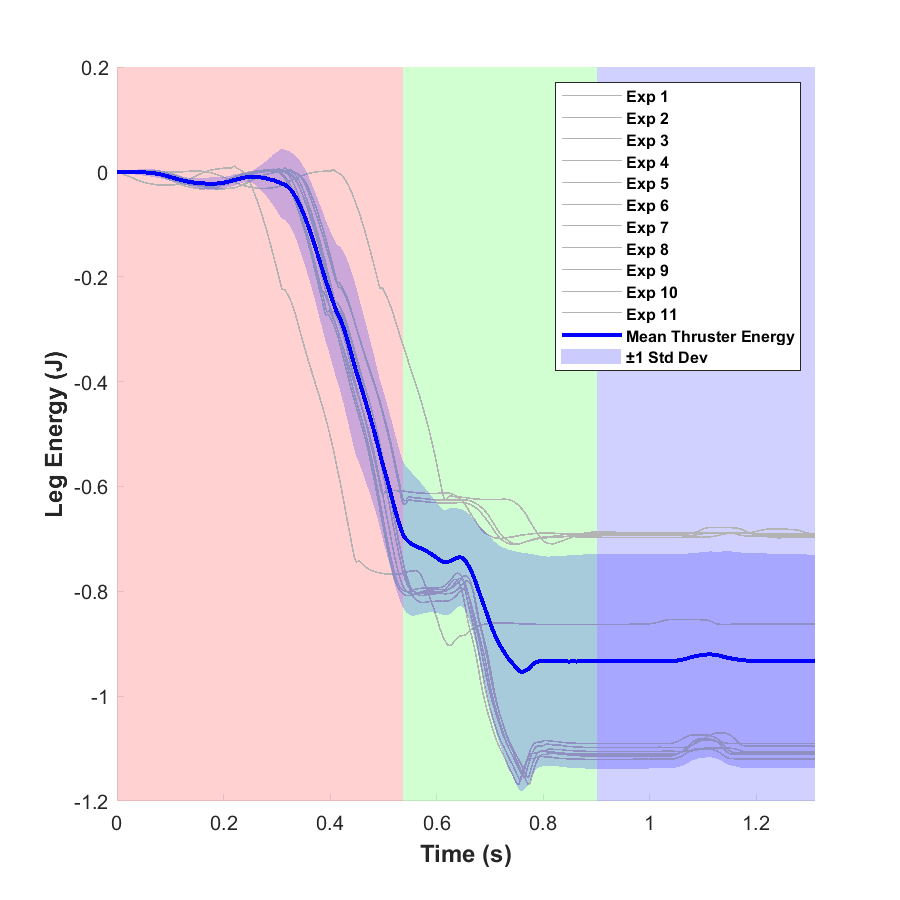}
  \caption{Illustrates Leg energy evolution during the jumping motion as mean and standard deviation plot of $11$ consecutive trails.}
  \label{fig:Jump-leg-energy}
\end{figure}
Figure~\ref{fig:Jump-leg-energy} presents the leg energy contributions during thruster-assisted jumping, calculated using the power formulation where for each joint, the instantaneous power is given by $P_j = \tau_j \, \dot{q}_j$ and the total energy contribution is $E_j = K_i \, \int_i \dot{q}_j$, where $\tau_j$ represents joint torque, $\dot{q}_j$ is the joint angular velocity, and $K_i$ is a normalization constant. The joint torque is obtained from current using $K_t \, i_{t}$, where $K_t$ is the torque constant obtained by performing joint characterization. The energy profiles from eleven experimental trials reveal critical insights into the hybrid propulsion strategy employed by the bipedal robot. During the take-off phase $(0-0.4s, \text{red region})$, the leg joints collectively generate approximately $-0.8$ J of energy, with the rapid energy decrease indicating active power generation for vertical propulsion. This substantial leg contribution confirms that despite thruster assistance, leg motion remains the dominant source of propulsion during jump initiation. The steepest energy gradient occurs between $0.3-0.5$s, coinciding with the transition from take-off to ballistic flight, where the legs must generate maximum power to achieve lift-off velocity. During the ballistic phase $(0.4-0.8s, \text{green region})$, the total leg energy reaches a minimum of approximately $-1.1$ J, representing the cumulative work done by all leg joints to achieve the jump. The consistency across multiple trials, evidenced by the tight standard deviation bands, demonstrates robust and repeatable energy management strategies. The mean thruster energy contribution (bold blue line) provides a reference for comparing propulsive contributions, though the dominant leg energy magnitudes during take-off underscore that this remains fundamentally a leg-driven jump enhanced by thrusters rather than a thruster-dominated flight. These energy profiles validate the hybrid nature of the system while emphasizing that effective leg joint coordination remains essential for successful jumping performance even with thruster augmentation.
\begin{figure}[H]
  \centering
    \includegraphics[width=1\textwidth]{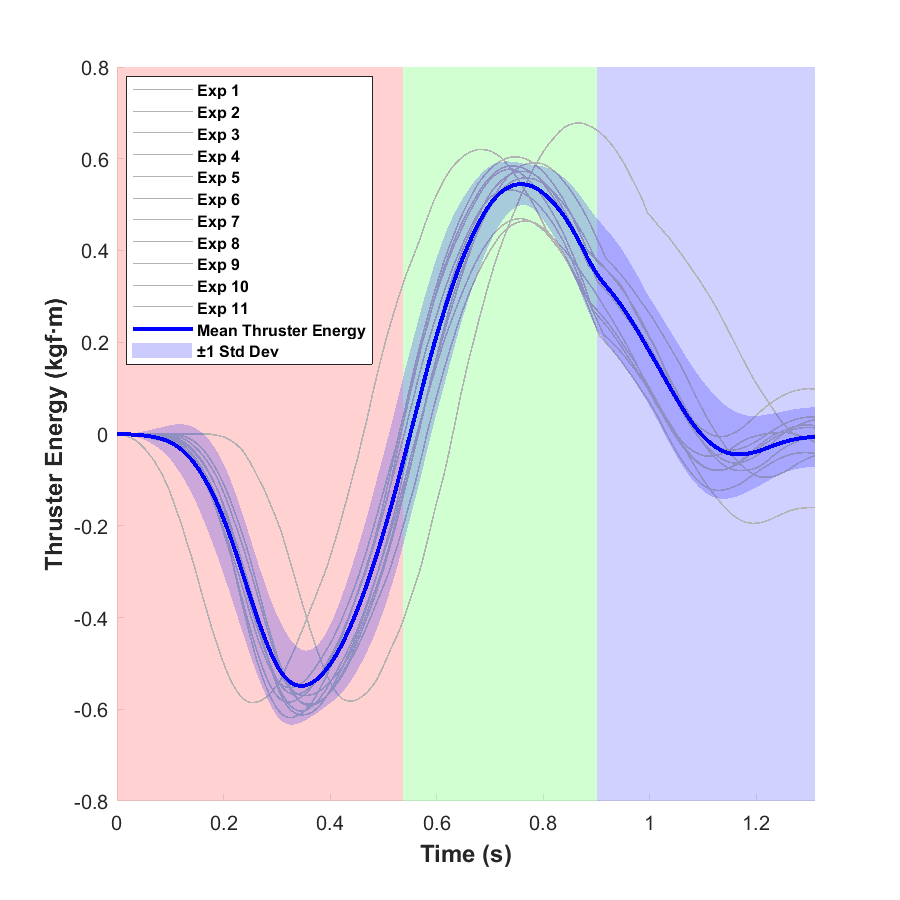}
  \caption{Shows the energy generated by the thruster during 11 trails of jump motion.}
  \label{fig:Jump-thrust-energy}
\end{figure}
Figure~\ref{fig:Jump-thrust-energy} illustrates the thruster energy contribution throughout the jumping motion, calculated using the formulation where the vertical thrust force is $F_z = F_{tot,mag} \sin(\theta_p)$ and instantaneous power is $P_z = F_z \, v_z$, where $F_{tot,mag}$ represents the total thrust magnitude, $\theta_p$ is the body pitch angle obtained from the EKF estimator, and $v_z$ is the vertical velocity of the torso.The thruster energy profile reveals a characteristic pattern that complements the leg-dominated propulsion strategy observed in the previous analysis. During the take-off phase, the thruster energy decreases from $0$ to approximately $-0.6$ kgf·m, indicating active thrust generation that augments the leg propulsion. The relatively smooth energy decrease suggests well-coordinated thruster activation that ramps up gradually to avoid destabilizing the robot during the critical leg push-off period. The minimum energy point coincides with the take-off to ballistic transition, where maximum thrust power is delivered to enhance the jump height. During the ballistic phase, the thruster energy increases to a maximum of approximately $0.6$ kgf·m, representing the total work done by the thrusters to extend flight time and maintain stability. The energy plateau during mid-flight indicates sustained thrust output, while the gradual increase suggests either thrust magnitude reduction or unfavorable pitch angles that reduce the vertical thrust component according to the $\sin(\theta_p)$ relationship. Notably, the thruster contribution during the ballistic phase is crucial for extending flight duration beyond what pure leg propulsion would achieve. The touch-down phase shows thruster energy returning to near zero, confirming that thrusters are deactivated during landing to allow the legs to handle impact absorption naturally. This strategic thruster shutdown prevents interference with the leg compliance mechanisms essential for safe landing. The consistency across eleven experimental trials, shown by the tight standard deviation bands, demonstrates reliable thruster control implementation. Comparing the thruster energy magnitudes $±0.6$ kgf.m with the previously shown leg energy contributions $±1.1$ J, it becomes evident that while thrusters provide valuable assistance, the legs remain the dominant source of propulsion and impact absorption during lift-off and touch-down phases respectively. This energy analysis confirms the hybrid system's design philosophy where thrusters enhance rather than replace bipedal jumping capabilities, with their primary benefit being extended flight time and improved aerial stability rather than fundamental propulsion generation.
\begin{figure}[H]
  \centering
    \includegraphics[width=1\textwidth]{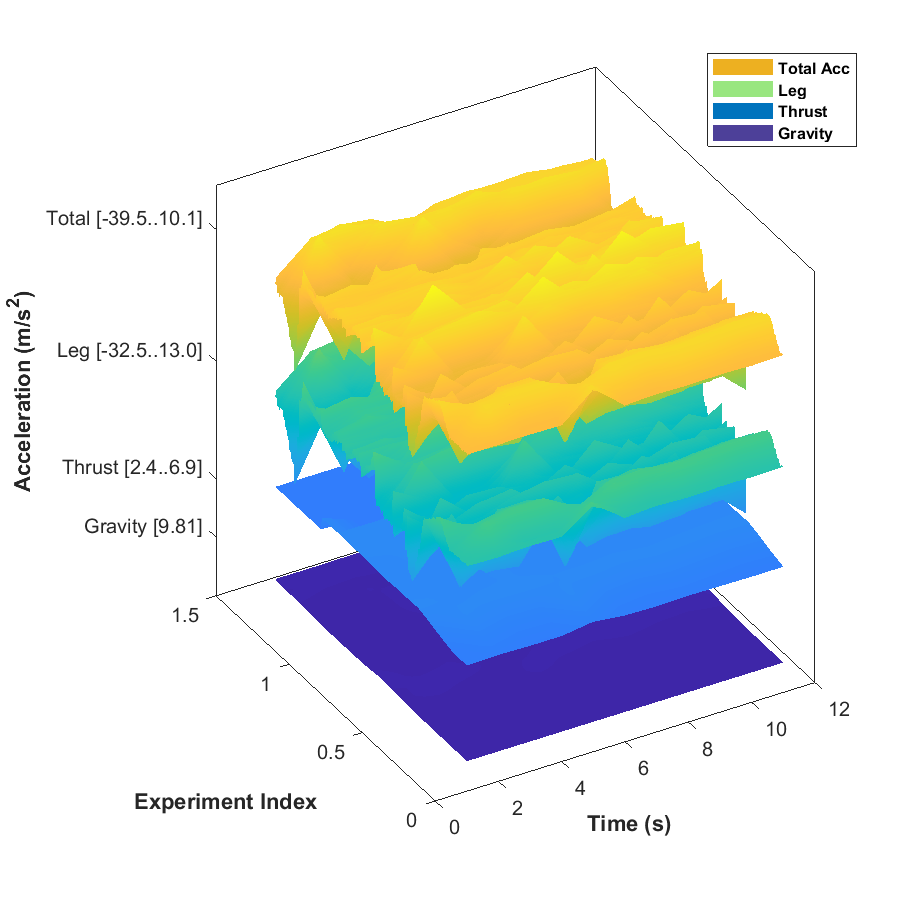}
  \caption{Illustrated the evolution body z acceleration (yellow) which is further distributed into individual components i.e. gravity (purple), Thruster (blue), Leg (green).}
  \label{fig:Jump-acc}
\end{figure}
Figure~\ref{fig:Jump-acc} presents a comprehensive analysis of acceleration components during thruster-assisted bipedal jumping across multiple experiments, decomposing the total acceleration measured by the IMU into its constituent sources. The total acceleration $a_{\text{tot}} = \ddot{p}_{B,z}$ is decomposed according to
\begin{align}
    a_{\text{tot}} &= a_{\text{uth}} + a_l + a_g \\
\end{align}
where $a_g = g = -9.8 \text{ m/s}^2$ represents gravitational acceleration (constant). Further, thruster induced accelertion is obtained as $a_{\text{uth}} = \frac{F_{\text{mag}} \sin(\theta_p)}{m_B}$, where $F_z = F_{\text{mag}} \sin(\theta_p)$ is the thruster force in the z direction and $m_B = 6.5$kg is the mass of robot. Lastly, leg acceleration is obtained using $a_l = a_{\text{tot}} - a_{\text{uth}} - a_g$. The 3D visualization reveals distinct acceleration patterns across the jump phases for all experiments. The gravity component (purple) maintains a constant $-9.81 \frac{\text{m}}{s}^{2}$ throughout all trials, serving as the baseline against which active accelerations must work. During the take-off phase, the leg component (green) generates substantial upward accelerations ranging from $-32.5$ to $-13.0$ m/s², demonstrating the dominant role of leg actuation in jump initiation. The negative values indicate upward acceleration opposing gravity, with peak leg accelerations occurring around 0.3-0.4s when maximum leg extension power is generated. The thruster component (blue) shows more modest contributions of $2.4$ to $6.9$ $\frac{\text{m}}{s}^{2}$, activated primarily during late take-off and throughout the ballistic phase. The temporal evolution shows that thrusters complement rather than dominate the acceleration profile, with their contribution becoming more significant during the flight phase when legs can no longer generate force. The total acceleration (yellow) reaches peak values of $-39.5$ to $-10.1$ $\frac{\text{m}}{s}^{2}$ during take-off, representing the combined effect of all sources. Notably, the total acceleration surface shows considerable variation across experiments, suggesting differences in jump execution strategies or thruster timing. During the ballistic phase $(0.5-1.0s)$, the leg acceleration drops to near zero as expected, leaving only thruster and gravitational accelerations active. The thruster contribution during this phase is crucial for extending flight time, maintaining accelerations that partially counteract gravity. The consistency of the decomposition across trials validates the measurement and calculation methodology, while the variations in magnitude highlight the system's flexibility in achieving jumping motion through different combinations of leg and thruster contributions. This acceleration analysis quantitatively confirms that successful jumping in this hybrid system relies primarily on leg-generated accelerations for take-off, with thrusters providing supplementary lift and flight extension capabilities.
\begin{figure}[H]
  \centering
    \includegraphics[width=1\textwidth]{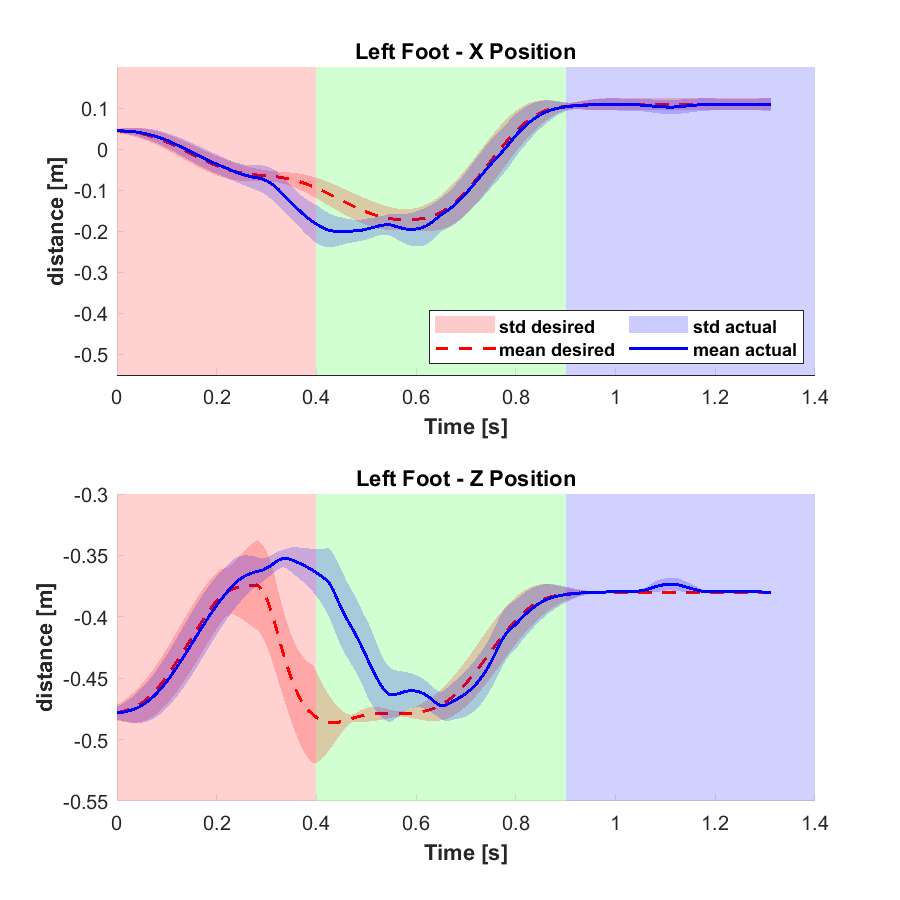}
  \caption{Shows the desired and actual foot position in body frame.}
  \label{fig:Jump-foot-pos}
\end{figure}
Figure~\ref{fig:Jump-foot-pos} compares the desired and actual foot trajectories for the left foot during thruster-assisted jumping, revealing the tracking performance of the robot's control system across different jump phases. The foot position are calculated from the forward kinematics using the desired and actual joint angles. In the horizontal ($x$) direction, the foot begins at approximately $0.05$ m and moves backward to $-0.2$ m during the take-off phase (red region, $0$--$0.4$ s), with the actual trajectory closely following the desired path indicated by the red dashed line. The most significant tracking deviation occurs during the late take-off phase around $0.4$--$0.5$ s, where the actual foot position shows oscillations not present in the smooth desired trajectory, likely due to the dynamic effects of thrust initiation and ground reaction forces. During the ballistic phase (green region, $0.4$--$0.8$ s), both trajectories converge as the foot swings forward from $-0.2$ m to $0.1$ m, preparing for landing configuration. The close alignment during flight suggests effective trajectory planning that accounts for the natural leg dynamics during the aerial phase. In the vertical ($z$) direction, the foot height varies from $-0.5$ m to $-0.35$ m throughout the motion, with the trajectory showing characteristic double-peak behavior. The first peak at approximately $0.3$ s corresponds to the leg extension during take-off, while the second peak at $0.9$ s occurs during the landing preparation phase. Notable tracking errors appear during the transition between take-off and ballistic phases ($0.4$--$0.6$ s), where the desired trajectory maintains a constant height of $-0.48$ m while the actual foot position continues to rise to $-0.45$ m. This deviation suggests that the controller prioritizes stability over precise tracking during this critical phase transition. The touch-down phase (blue region, $1.0$--$1.4$ s) shows excellent trajectory convergence in both directions, with the foot successfully reaching and maintaining the target landing position of $(0.1, -0.38)$ m. The standard deviation bands indicate consistent performance across multiple trials, with larger variations during the dynamic take-off phase and tighter bounds during the controlled landing phase. Overall, the tracking performance demonstrates that while minor deviations occur during high-acceleration phases, the control system successfully guides the foot through the essential waypoints required for stable jumping and landing in this hybrid locomotion system.
\begin{figure}[H]
  \centering
    \includegraphics[width=1\textwidth]{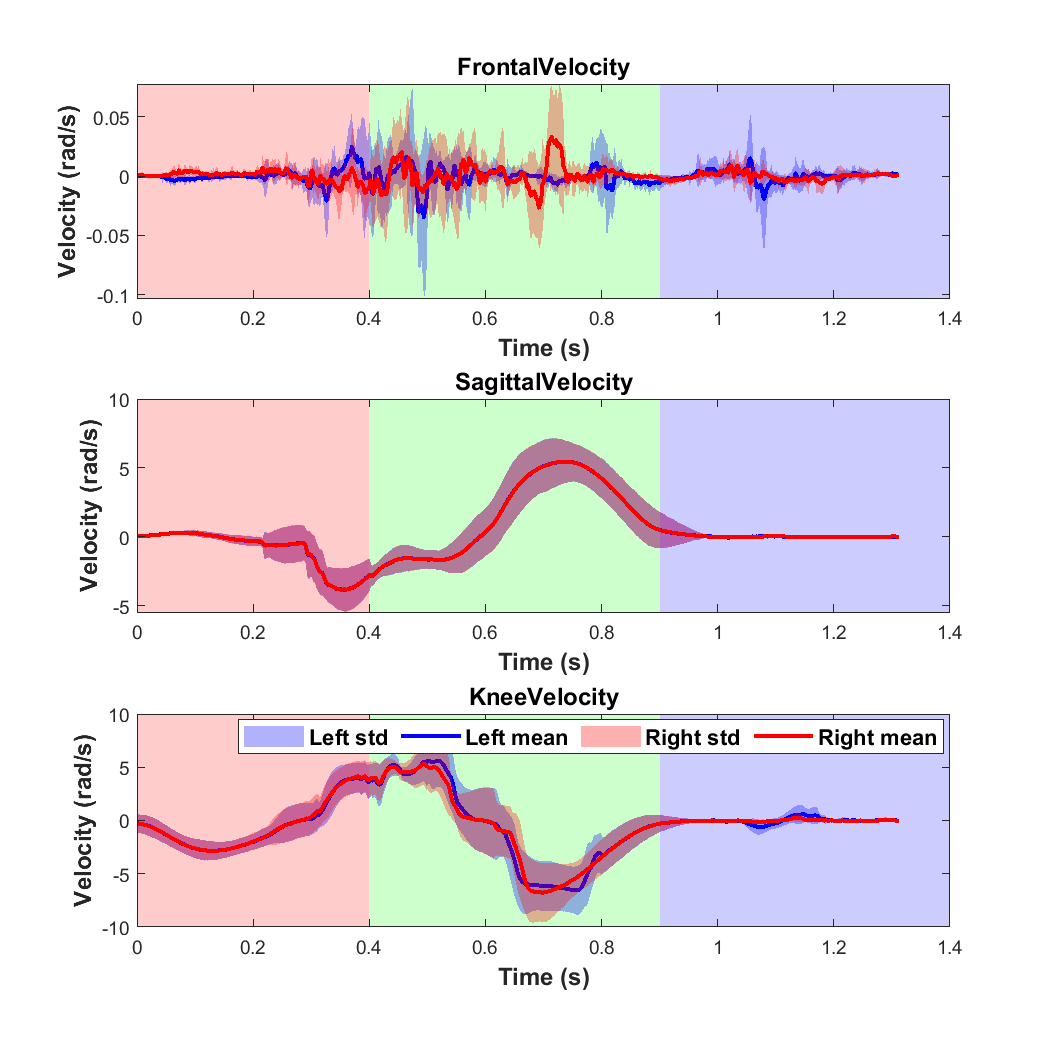}
  \caption{Illustrates the joint velocity for frontal, saggital and knee for both foot.}
  \label{fig:Jump-joint-velo}
\end{figure}
Figure~\ref{fig:Jump-joint-velo} presents the angular velocity profiles for the frontal, sagittal, and knee joints during thruster-assisted jumping, revealing distinct motion patterns that characterize each phase of the hybrid locomotion. The frontal plane velocities remain near zero throughout the entire jump cycle, with fluctuations within $\pm 0.05$ rad/s, confirming minimal lateral motion and good stability during jumping. The small oscillations during the ballistic phase (green region, $0.4$--$0.8$ s) likely result from thruster-induced vibrations or minor asymmetries in thrust application, though these perturbations remain well-controlled. The sagittal joint velocities show a characteristic bi-phasic pattern essential for forward momentum generation and balance control. During take-off ($0$--$0.4$ s), sagittal velocities decrease from $0$ to approximately $-3$ rad/s, indicating backward rotation that helps position the center of mass over the support base. This is followed by a rapid reversal to positive velocities reaching $+6$ rad/s during the take-off to ballistic transition, facilitating forward body rotation that coordinates with the jump trajectory. The gradual return to zero velocity during touch-down ($1.0$--$1.4$ s) demonstrates controlled landing preparation. The knee joint velocities exhibit the most pronounced changes, reflecting their primary role in jump propulsion and landing absorption. During early take-off ($0$--$0.3$ s), knee velocities gradually increase as the joints prepare for explosive extension. The peak extension velocity of approximately $5$ rad/s occurs at $0.4$--$0.5$s, coinciding with maximum power generation for vertical propulsion. The subsequent negative velocities reaching $-5$ rad/s during late ballistic phase indicate knee flexion in preparation for landing, with the timing of this reversal critical for proper foot placement. During touch-down, knee velocities rapidly return to zero as the joints absorb impact energy and stabilize the robot. The close correspondence between left and right leg velocities across all joints demonstrates excellent bilateral symmetry in the control strategy, essential for maintaining balance during this dynamic maneuver. The tight standard deviation bands, particularly for knee and sagittal joints, indicate consistent execution across multiple trials, while the slightly larger variations in frontal velocities reflect the system's active compensation for minor lateral disturbances.
\begin{figure}[H]
  \centering
    \includegraphics[width=1\textwidth]{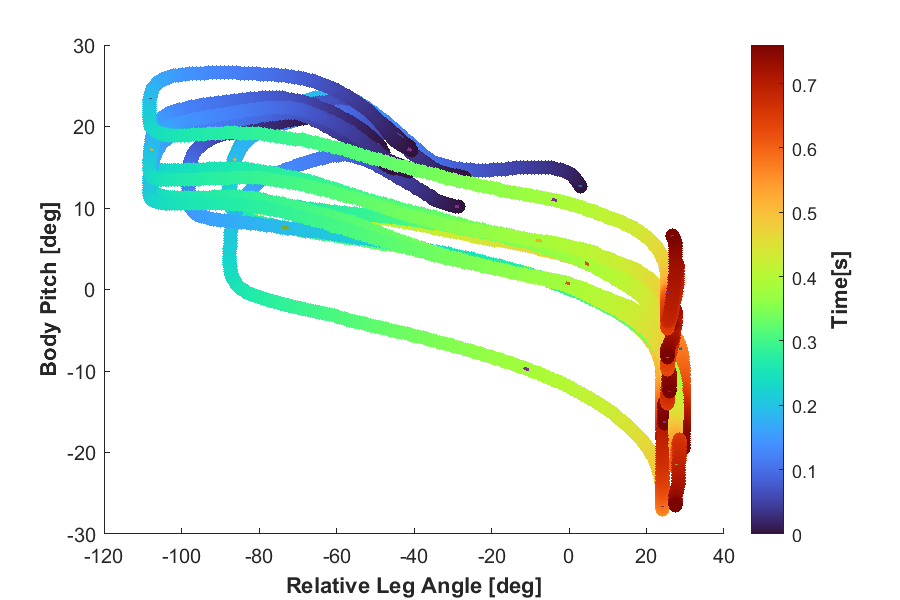}
  \caption{Illustrates relationship between left leg sagittal joint and body pitch angle.}
  \label{fig:Jump-ang-phase}
\end{figure}
Figure~\ref{fig:Jump-ang-phase} presents a phase portrait of body pitch angle versus relative leg angle throughout the jumping motion, with the color gradient representing temporal evolution from jump initiation (dark blue, $t = 0$ s) to landing completion (dark red, $t \approx 0.7$ s). This phase space representation reveals the complex coordination between leg configuration and body orientation required for successful thruster-assisted jumping. The trajectory begins at a relative leg angle of approximately $20 \deg$ with near-zero body pitch, representing the initial standing configuration. During the early take-off phase (dark blue to light blue, $0$--$0.2$ s), the system follows a counterclockwise trajectory as the legs extend (increasing relative angle to $\sim 40 \deg$) while maintaining relatively stable body pitch between $20 \deg$ and $25 \deg$. This initial phase demonstrates the robot's ability to generate leg extension without inducing excessive body rotation, crucial for vertical jump initiation. The most dramatic phase space evolution occurs during the take-off to ballistic transition (light blue to green, $0.2$--$0.4$ s), where the trajectory rapidly traverses from positive to negative relative leg angles while body pitch decreases from $25 \deg$ to approximately $-5 \deg$. This transition represents the critical moment when the legs complete their extension and begin to retract for flight configuration, while the body simultaneously pitches forward to align with the desired flight trajectory. During the ballistic phase (green to yellow, $0.4$--$0.5$ s), the system exhibits a tight clustering of states around $-60 \deg$ to $-80 \deg$ relative leg angle and $0 \deg$ to $10 \deg$ body pitch, indicating a stable flight configuration where the legs are retracted and body orientation is controlled by thruster modulation. The landing preparation and touch-down phase (yellow to dark red, $0.5$--$0.7$ s) shows a rapid transition to highly negative body pitch angles (reaching $-25 \deg$) while the relative leg angle moves toward $30°$, representing leg extension for landing while the body pitches backward to prepare for impact absorption. The phase portrait's loop-like structure indicates successful completion of the jump cycle, though the ending configuration differs from the starting point, suggesting either incomplete recovery to standing position within the time window shown or adaptation to post-landing stability requirements.

% conclusion and Future work
 \chapter{Conclusion}
\label{chap:conclusion}

This thesis investigated the coordinated control of legged robotic locomotion, focusing on hybrid trotting and jumping behaviors underactuated along certain degrees of freedom. Through real-world experiments, simulation analysis, and control evaluation, we established that stable and robust locomotion is achievable across dynamic motion phases.

We first demonstrated that the system maintains bounded trajectories and consistent foot placement, confirming gait stability. The synergy between leg and thruster modules was evident: while the legs handled propulsion and impact absorption, thrusters played a critical stabilizing role during the ballistic phase, leading to smoother aerial trajectories and controlled landings.

Joint-level analysis showed that torque demands remained low, and symmetry across joint tracking was preserved. This further validates our efficient control architecture, which leverages natural dynamics to minimize actuation effort. Additionally, foot placement stayed within kinematic limits, with only minor deviations at phase transitions, indicating good robustness and tracking performance.

Importantly, UDoF stability was confirmed. Visualizations via stick diagrams and limit cycles revealed bounded oscillations and no drift, supporting the emergence of stable limit cycles in hybrid leg–thruster systems.

A key observation was the strong nonlinear coupling between body and leg motion during the ballistic phase. Specifically, leg displacement was up to 4× greater than body displacement, and we identified a noticeable asymmetry between take-off and touch-down phases. This asymmetry highlights an unsolved challenge: a unified controller may not be sufficient across all motion phases.

\subsection{Learnings from the Analysis}
The experimental results on both trotting and jumping dynamics of Harpy yield several important learnings:

\begin{itemize}
    \item \textbf{Effectiveness of the gantry setup:} The lightweight gantry successfully constrained lateral and rotational dynamics, allowing clear isolation of sagittal-plane motion. Minor lateral deviations and COM oscillations confirmed that the setup imposed minimal disturbance, validating it as a reliable testbed for controlled locomotion studies.  

    \item \textbf{Body position and orientation tracking:} During trotting, Harpy maintained bounded pitch ($\pm 5^{\circ}$) and roll/yaw ($\pm 10^{\circ}$), confirming stable body orientation. The EKF-based fusion of IMU and OptiTrack data provided smooth and reliable state estimates, essential for controller validation.  

    \item \textbf{Foot placement repeatability:} Parameterized foot trajectories highlighted consistent step patterns in trotting, with systematic forward/backward reach and vertical excursion. In, jumping experiments, the thrusters help in reducing the tracking error.

    \item \textbf{Underactuated stability:} Limit cycle analyses across body positions and joint states demonstrated bounded and repeatable oscillations. The existence of closed trajectories in phase space confirms that Harpy exploits passive dynamics for stability for a robust underactuated locomotion.  

    \item \textbf{Joint behavior and tracking:} Trotting revealed increasing error variance in frontal and sagittal joints but stable knee tracking, reinforcing the knee’s role in impact absorption and load consistency. Jumping torque and velocity profiles confirmed that legs remain critical in initiating propulsion, controlling flight configuration, and stabilizing landing despite thruster assistance.  

    \item \textbf{Energy contributions of legs and thrusters:} Energy decomposition showed that legs dominate take-off propulsion, with thrusters providing supplementary lift and extended aerial stability. Thrusters were deliberately deactivated at landing, enabling natural compliance of the legs to absorb impact. This validates Harpy’s hybrid design philosophy: thrusters enhance rather than replace leg actuation.  

    \item \textbf{Leg–body coordination:} Phase portraits of body pitch versus relative leg angle illustrated the intricate coupling between leg motion and body orientation. Leg retraction during ballistic flight consistently helped counter pitch variations that thrusters could not control because the thrusters cannot contribute to the pitch movement since they are located close to the COM.  
\end{itemize}

\subsection{Final Discussion on Results}
The combined analysis of trotting and jumping experiments demonstrates that Harpy achieves stable and repeatable locomotion under gantry-constrained conditions. The results validate both the hardware design and control framework while revealing the distinct yet complementary roles of legs and thrusters.  

For trotting, the robot exhibited reliable gait regularity, bounded orientation, and stable underactuated dynamics, confirming that the locomotion system leverages passive dynamics while maintaining effective control. The observed joint-space limit cycles further highlighted energy-efficient motion patterns consistent with natural locomotion strategies.  

For jumping, the experiments emphasized the hybrid nature of Harpy’s design. Legs were confirmed as the primary source of propulsion during take-off and impact absorption during landing, while thrusters extended flight time and stabilized body orientation during the ballistic phase. The coupling between leg dynamics and body orientation played a critical role in mitigating pitch instabilities that thrusters could not counteract due to their torso-mounted configuration.  

Overall, the findings reinforce the utility of the gantry-assisted setup for isolating sagittal dynamics and provide strong evidence that Harpy’s hybrid actuation strategy is both functional and robust. These insights directly inform the next stage of controller development: designing adaptive strategies for improved landing consistency, optimizing energy distribution between legs and thrusters, and refining feedback control to handle variability in high-impact phases. Such enhancements will be essential for transitioning Harpy from constrained experiments toward unconstrained, real-world locomotion.

\section{Future work}

The observed asymmetry between the take-off and touch-down phases suggests that a uniform control strategy may not be sufficient to ensure robust performance throughout the entire jump cycle. As a potential future direction, phase-specific controllers can be introduced to better handle the distinct dynamics of each phase. Event-based PID controllers could provide simple yet effective regulation around critical transitions, while cascaded MPC offers a more systematic framework for optimizing performance under constraints across multiple time horizons. Alternatively, cascaded NLP approaches can be explored to capture the full non-linear dynamics of the system, enabling more precise and energy-efficient jump execution. Furthermore, future work should also focus on removing the current gantry setup, which was primarily used to isolate and study sagittal-plane dynamics, in order to test the robot’s performance in fully unconstrained conditions and evaluate the role of frontal-plane dynamics during trotting and jumping.

% --- Bibliography ----
\bibliographystyle{IEEEtran}  %'plain' for standard, 'unsrt' for correct order

% include bibliography definition
\bibliography{references-3,references,references-2}

% --- Appendix ---
\appendix
%include anything you need in the appendix
%\include{tex/appendixA}

% --- Index ----
%\printindex

% --- that's it ---
\end{document}